\documentclass[12pt,draftclsnofoot,onecolumn]{IEEEtran}

\pdfoutput=1

\usepackage{mathrsfs}
\usepackage{cite} 
\usepackage{xspace}
\usepackage{filecontents}
\usepackage{float}
\usepackage{pgf, tikz}
\usepackage{graphicx}
\usepackage[cmex10]{amsmath}
\usepackage{mathtools} 

\usepackage{bm}
\usepackage{amssymb} 
\usepackage{array,algorithm,algorithmic}
\usepackage{amsfonts}
\usepackage{amsthm}   
\usepackage{enumerate} 
\usepackage{enumitem}
\newtheorem{definition}{Definition} 
\newtheorem{theorem}{Theorem} 
\newtheorem{lemma}{Lemma}
\newtheorem{proposition}{Proposition}
\newtheorem{remark}{Remark}

\usepackage{comment}

\usepackage{stfloats,epstopdf}
\usepackage{enumerate}
\usepackage{pgfplots}
\usepackage{caption}
\usepackage{subcaption}
\usepackage{tabularx}
\usepackage{pgfplotstable}
\usepgfplotslibrary{dateplot}
\usepackage{cite}
\usepackage{url}
\usepackage{amsbsy}
\usepgfplotslibrary{patchplots}
\usepackage{xcolor}
\usepackage{tikz}
\usetikzlibrary{shadows}
\usetikzlibrary{fadings}
\usetikzlibrary{arrows}
\usetikzlibrary{shapes,snakes}
\pgfplotsset{compat=1.7}
\usetikzlibrary{patterns}
\usetikzlibrary{spy}

\usepackage{amsmath,accents}
\DeclareMathSymbol{\widehatsym}{\mathord}{largesymbols}{"62}

\DeclareMathSymbol{\widetildesym}{\mathord}{largesymbols}{"65}

\pgfplotscreateplotcyclelist{mycyclelist_no_marks}{
solid, red, line width=1.0pt, every mark/.append style={solid, fill=red}, mark=none\\%
densely dashed, blue, line width=1.0pt, every mark/.append style={solid, fill=blue}, mark=none\\%
}

\pgfplotscreateplotcyclelist{mycyclelist}{
solid, red, line width=1.0pt, every mark/.append style={solid, fill=red}, mark=square, mark size={1}\\%
densely dashed, blue, line width=1.0pt, every mark/.append style={solid, fill=blue}, mark=triangle, mark size={1}\\%
densely dotted, line width=1.0pt, green!70!black, every mark/.append style={solid, fill=green!70!black}, mark=o, mark size={1}\\%
densely dashdotted, line width=1.0pt, brown!70!black, every mark/.append style={solid, fill=brown!70!black}, mark=diamond, mark size={1}\\%
dotted, line width=1.0pt, orange!70!black, every mark/.append style={solid, fill=orange!70!black}, mark=x, mark size={1.3}\\%
thick, solid,  red\\%
densely dashdotted, line width=1.0pt, blue\\%
densely dotted,  line width=1.2pt, green!70!black\\%
solid,  blue, every mark/.append style={solid, fill=blue}, mark=*, mark size={1}\\%
dashed, blue, every mark/.append style={solid, fill=blue}, mark=*, mark size={1}\\%
solid,  red, every mark/.append style={solid, fill=red}, mark=square*, mark size={1}\\%
dashed, red, every mark/.append style={solid, fill=red}, mark=square*, mark size={1}\\%
solid,  green!60!black, every mark/.append style={solid, fill=green!60!black}, mark=triangle*, mark size={1.2}\\%
dashed, green!60!black, every mark/.append style={solid, fill=green!60!black}, mark=triangle*, mark size={1.2}\\%
solid,  brown!70!black, every mark/.append style={solid, fill=brown!70!black}, mark=diamond*, mark size={1.2}\\%
dashed, brown!70!black, every mark/.append style={solid, fill=brown!70!black}, mark=diamond*, mark size={1.2}\\%
}
\pgfplotscreateplotcyclelist{mycyclelist2}{
solid, red, every mark/.append style={solid, fill=red}, mark=square, mark size={1}\\%
densely dashed, blue, line width=1.0pt, every mark/.append style={solid, fill=blue}, mark=triangle, mark size={1}\\%
densely dotted, line width=1.0pt, green!70!black, every mark/.append style={solid, fill=green!70!black}, mark=o, mark size={1.2}\\%
densely dashdotted, line width=1.0pt, brown!70!black, every mark/.append style={solid, fill=brown!70!black}, mark=x, mark size={1.2}\\%
thick, solid,  red\\%
densely dashdotted, line width=1.0pt, blue\\%
thick, solid,  green!70!black\\%
dashed, line width=1.0pt, brown!70!black\\
densely dashdotted, blue, every mark/.append style={solid, fill=blue}, mark=square, mark size={1}\\%
solid, blue, every mark/.append style={solid, fill=blue}, mark=triangle, mark size={1.2}\\%
densely dotted, line width=0.5pt, blue, every mark/.append style={solid, fill=blue}, mark=o, mark size={1}\\%
dashed, blue, every mark/.append style={solid, fill=blue}, mark=x, mark size={1.2}\\%
thick, dashed,  green!70!black\\%
densely dashdotted, red, every mark/.append style={solid, fill=red}, mark=square, mark size={1}\\%
solid, red, every mark/.append style={solid, fill=red}, mark=triangle, mark size={1.2}\\%
densely dotted, red, every mark/.append style={solid, fill=red}, mark=o, mark size={1}\\%
dashed, red, every mark/.append style={solid, fill=red}, mark=x, mark size={1.2}\\%
}
\pgfplotscreateplotcyclelist{mycyclelist3}{
solid,  blue, every mark/.append style={solid, fill=blue}, mark=*, mark size={1}\\%
densely dashed, red, every mark/.append style={solid, fill=red}, mark=x, mark size={1.5}\\%
}

\newcommand{\black}{\color{black}}
\newcommand{\blue}{\color{blue}}
\newcommand{\red}{\color{red}}

\newcommand{\AuthorOne}{Sheeraz~A.~Alvi,~\textit{Member,~IEEE}} 
\newcommand{\AuthorTwo}{Yi~Hong,~\textit{Senior~Member,~IEEE}}
\newcommand{\AuthorThree}{Salman~Durrani,~\textit{Senior~Member,~IEEE}}
\newcommand{\ThankOne}{Sheeraz Alvi and Salman Durrani are with the School of Engineering, the Australian National University, Canberra, ACT 2601, Australia
	(emails: \{sheeraz.alvi, salman.durrani\}@anu.edu.au).
	
Yi Hong is with the Department of Electrical and Computer Systems Engineering, Monash University, Clayton, VIC 3800, Australia (email: yi.hong@monash.edu).

Corresponding author: Sheeraz A. Alvi.
}

\begin{document}
%
\title{Utility Fairness for the Differentially Private Federated Learning}
\author{\IEEEauthorblockN{\AuthorOne,~\AuthorTwo,\\~and~\AuthorThree\thanks{\ThankOne}}
}

\maketitle

\begin{abstract}
%
%
Federated learning (FL) allows predictive model training on the sensed data in a wireless Internet of things (IoT) network evading data collection cost in terms of energy, time, and privacy.
%
%
In this paper, for a FL setting, we model the learning gain achieved by an IoT device against its participation cost as its utility. The local model quality and the associated cost differs from device to device due to the device-heterogeneity which could be time-varying. We identify that this results in utility unfairness because the same global model is shared among the devices.
%
%
In the vanilla FL setting, the master is unaware of devices' local model computation and transmission costs, thus it is unable to address the utility unfairness problem. In addition, a device may exploit this lack of knowledge at the master to intentionally reduce its expenditure and thereby boost its utility. 
%
%
We propose to control the quality of the global model shared with the devices, in each round, based on their contribution and expenditure. This is achieved by employing differential privacy to curtail global model divulgence based on the learning contribution. Furthermore, we devise adaptive computation and transmission policies for each device to control its expenditure in order to mitigate utility unfairness.
%
%
Our results show that the proposed scheme reduces the standard deviation of the energy cost of devices by 99\% in comparison to the benchmark scheme, while the standard deviation of the training loss of devices varies around 0.103.
\end{abstract}

\IEEEpeerreviewmaketitle


\vspace{-0.1cm}

\section{Introduction}



The Internet of things (IoT) is largely based on the machine-type communication devices (MTDs) with sensing, computing, communication, and/or control capabilities, such as wearables, telemetry devices, surveillance devices, smartphones, etc., \cite{fuqaha-2015,dawy-2017}. Applying machine learning techniques on the sensing data acquired by these devices holds great potential of providing intelligent and personalized services to the user through predictive models \cite{9072101, 8270639}. However, this data is mostly private in nature, thereby collecting and storing this sensitive physical information in a centralized cloud server for learning is risky. In that, various scenarios exist where both parties may or may not be trustworthy. Moreover, due to the wireless and autonomous operation, these devices are mostly battery operated, thus the data communication is severely energy-constrained \cite{tcom-sheeraz, hanzo-2017}. 


To mitigate the cost of collecting sensed data (in terms of energy, time, privacy) at a central entity, the training can be performed locally at the devices. Clearly the trained models will perform poorly due to the limited data size. Therefore, a specialized machine learning setting referred to as the Federated learning (FL) provides a distributed learning methodology to learn a global model at a central entity without the need to access distributive data \cite{FL-1}. In FL, a master device iteratively learns the global model parameters by combining the updates of local model parameters computed and then shared by the participating devices with the master.
In a wireless IoT network, FL realizes effective predictive model learning with profound communication efficiency and strong privacy. From the master’s perspective, FL enables global model learning in a privacy-preserving manner, i.e., without collecting sensitive data from the MTDs. On the other hand, each MTD procures a global model which is learned using disjoint data of multiple MTDs, thus a more effective model as compared to its local model. We refer to this improvement achieved in the model learning against the accumulative cost associated with local model computation and transmission towards the master as \textit{MTD's utility}.



In the vanilla FL setting, each device receives the same global model from the master irrespective of the quality of its local model and the associated cost of computation and transmission. However, in a practical IoT network, the local model quality and accumulative cost vary between devices due to the (time-varying) diversity in the training data, wireless channel, computation/communication resources, etc. Furthermore, a malicious device may intentionally reduce the cost of computation and/or transmission, e.g., by decreasing the training dataset size, performing fewer iterations, re-transmitting the previous update, employing quantization/compression, etc. Thereby, the device increases its utility while enjoying the same global model as other legitimate devices. 
This malicious behavior causes utility unfairness among devices and damages the quality of the global model which in turn increases the convergence time. In particular, if legitimate devices increase their computation budget to maximize the quality of the local model to help achieve better global model, their additional investment does not yield productivity gains due to the poor local models contributed by the malicious devices. Nevertheless, distinguishing this malicious behaviour from the natural channel/resource diversity is rather hard without additional information. In the existing FL settings, there are no standard of operations devised for this situation, thus the master cannot ensure pertinent operation of the devices.



Security and device heterogeneity are the two main issues for FL in a wireless IoT network setting. By evading data transmission, FL does protect devices from conventional eavesdropping attacks. Nevertheless, the private information can still be revealed by manipulating the transmitted local models. For example, a model inversion attack analyses the differences in the shared parameters (e.g., weights of a neural network) to reconstruct the training data of a device \cite{model_inversion, 9048613}. Similarly, a membership inference attack can reveal if a data sample was used in the training process \cite{inference, 9069945}. Thus, the training data is somewhat vulnerable in a FL setting. In this regard, differential privacy (DP) has been proved to offer quantifiable protection against information leakage \cite{SnS-Dwork2014b, 1065184}. DP is a proactive method of adding artificial noise to the data before sharing (we will discuss it in greater detail in the later sections). Various existing studies proposed DP based learning algorithms. Local DP (LDP) is employed in \cite{8640266}, in which users send a randomized version of their local data for distributed estimation over user uploaded data. In \cite{9152691}, DP based Stochastic gradient descent algorithm is analysed considering different dataset sizes and privacy levels.

To improve user services, a DP based personalized FL method is proposed in \cite{personal} for a wireless IoT network. DP is just employed for protection and the focus is on the adaptive training which considers device heterogeneity and data ownership. In \cite{9253545}, DP based FL algorithm is proposed for Internet of Vehicles (IoV), which minimize the communication overhead while achieving high accuracy in a DP based secure manner. In \cite{9317806}, a DP based FL algorithm is employed to devise an incentive model based on computation, communication, and privacy cost of the devices. A higher cost yields a higher (unspecified) reward and the device's utility is the difference between the reward and the cost. Similar incentive mechanisms for FL are proposed in \cite{iot13, iot14, iot12, iot16}. From the perspective of the master's profit, some have evaluated device contribution based on the local training delay \cite{iot13, iot14}, others have evaluated device contribution based on the training dataset \cite{iot12, iot16}.

\vspace{-0.1cm}
\subsection{Paper Contributions}

Prior works on FL in a wireless IoT network only focused on the expenditure or contribution of the devices and proposed some (typically unspecified) reward/incentive in response to the device heterogeneity. A few have employed DP only for protection against attacks via eavesdropping. To the best of our knowledge, no prior work has jointly considered the diversity in the expenditure and contribution of the devices impacting the master's model learning, and addressed the unfairness among devices or the malicious behaviour mentioned above.

We consider a FL setting, in which multiple heterogeneous MTDs cooperate with a master. To address the utility unfairness problem among MTDs, we propose to control the quality of the global model shared with the MTDs, in each round, based on their contribution and expenditure. 
We design a utility function for MTDs to model the learning gain and cost associated with it in each round. In particular, the utility function works as a catalyst and it is used to reveal the optimal computation and transmission policies such that the learning gain versus the cost is similar for all MTDs. 
This is achieved by treating global model as a precious commodity and controlling its quality through DP. Accordingly, the master will add noise in the global model before sharing it with a MTD in proportion to the deviation of its local model from the global model. In the proposed methodology, the master relies only on the local model quality to decide the level of noise to be added in each round. It is because the computation and transmission costs cannot not be quantified effectively. We use DP in two-folds, first to provide security and second to curtail global model divulgence. Through an extensive performance analysis of the proposed scheme we seek answers to the following two questions:

\begin{enumerate}

\item How do heterogeneous MTDs achieve similar utility (learning gain versus cost) without compromising the learning rate of the global model?

\item How does a MTD adapt its computation and transmission parameters in each round without any knowledge of the contribution and expenditure of other MTDs?

\end{enumerate}

\noindent Our investigation leads to the following observations:

\begin{itemize}

\item Our results show that the existing FL in a wireless IoT network setting suffers from severe utility unfairness among MTDs. The proposed scheme addresses this problem by controlling the productivity gains of MTDs and achieves similar learning gain and energy expenditure across all participating MTDs.

\item For each round, the proposed scheme produces optimal computation and transmission policies for individual MTDs without any knowledge of the contribution and expenditure of other MTDs. Similarly, the master controls the global model quality without any knowledge of the expenditure of MTDs.

\item Our results show that the proposed scheme reduces the standard deviation of the energy cost across MTDs by 99\% in comparison to the benchmark scheme, while the standard deviation of the training loss across MTDs varies around 0.103. Also, the proposed scheme provides about 12.17\% reduction in the average energy cost of MTDs as compared to the benchmark scheme.

\end{itemize}

\textit{Paper organization:} The rest of the paper is organized as follows. The system model and operation are presented in Section~II. The proposed differentially-private FL problem is formulated and solved in Section~III. Simulation results are presented in Section~IV. Section~V concludes the paper.

\section{System Model and Operation}

In this section, we present the overall network setup and its operation including the underlying FL model, the differential privacy mechanism, the communication channel model, and lastly the computation and transmission energy cost models.

\vspace{-0.1cm}
\subsection{Network Setup}

We consider a network consisting of a access point (AP) serving a set $\mathcal{K}=\{1,2,\cdot\cdot\cdot,K\}$ of heterogeneous MTDs. The network is composed of a single cell with the AP located at the center and the MTDs are located at arbitrary distances from the AP. We assume the AP has ample computation and energy resources to facilitate the collaborative FL.
All MTDs possess sensing, training, and transmission capability. However, each MTD has different computing and energy constraints. Each MTD regularly acquires some physical information from the environment and stores it in its local database. The $k$-th MTD has a local training dataset $\mathcal{D}_k = ( \mathcal{D}_{k,1}, \mathcal{D}_{k,2}, \cdot\cdot\cdot , \mathcal{D}_{k,d_k}) \in \mathbb{R}^{d_k} = \{\mathbf{x}_{k,i} \in \mathbb{R}^s, y_{k,i} \in \mathbb{R}\}^{d_k}_{i=1}$, where $k \in \mathcal{K}$, $\mathbf{x}_{k,i}$ denote a feature vector, $y_{k,i}$ denote the corresponding label, $d_k = |\mathcal{D}_k|$ is the total number of training samples stored in $k$-th MTD's database and $|\cdot|$ denote the cardinality of a set. 
We assume the MTDs possess different finite storage capacity to store data samples. After each sensing interval a given number of old data samples are replaced by equal number of freshly acquired data samples. Therein, the dataset size is assumed to be initially at full capacity and remains the same throughout the training process. Each MTD performs local training over its dataset and transmits the specific training parameters to the AP within a time block of $T$ secs.


\vspace{-0.1cm}
\subsection{Federated Learning}

The overall objective of the system is to learn a statistical model at the AP, over the datasets of all  participating MTDs. Accordingly, the AP needs to find a fitting vector $\mathbf{w}_\textup{g} \in \mathbb{R}^v$ (a global model) which minimizes some loss function for the given datasets. This learning task is formulated as follows \cite{wang-2018}:
\begin{equation}\label{gl-opt-prob}
\begin{aligned}
& \underset{ \mathbf{w}_\textup{g} \in \mathbb{R}^v} {\textup{minimize}}
& & \mathcal{G}(\mathbf{w}_\textup{g}) = \frac{1}{d} \sum_{k=1}^{K} d_k \mathcal{L}_k(\mathbf{w}_\textup{g}),
\end{aligned}
\end{equation}
\noindent where $d =\sum_{k=1}^{K} {d_k}$ is the total number of training samples of all MTDs and $d_k$ is the number of training samples of the $k$-th MTD, $\mathcal{G}(\cdot)$ is the total loss function which measures the empirical loss of all training samples of all MTDs, and $\mathcal{L}_k(\cdot)$ is total loss function for the $k$-th MTD which is defined as
\begin{equation}\label{locallossfunc}
 \mathcal{L}_k(\mathbf{w}_\textup{g}) = \frac{1}{d_k} \sum_{i=1}^{d_k} \ell(\mathbf{w}_\textup{g},\mathbf{x}_{k,i}, y_{k,i}), 
\end{equation}
\noindent where $\ell(\cdot)$ is a convex loss function. The choice of the loss function depends on the learning task. The two main streams of loss functions are regression loss functions (e.g, mean-squared-error) and classification loss functions (e.g., binary cross-entropy, Kullback-Leibler divergence).

To solve problem in \eqref{gl-opt-prob}, the AP needs access to all datasets. This implies that all MTDs must transmit their individual datasets to the AP, these datasets can be large and thus may incur huge data transmission cost on the MTDs. In many IoT cases, the complete dataset transmission is infeasible for MTDs due to the power constraint. In addition, the data sensed by these MTDs may contain sensitive information, e.g., user activity, medical history, etc. To this end, FL allows global model training in a privacy-preserving manner, i.e., without acquiring local training datasets at a central entity. 

We employ a generic FL mechanism in \cite{konevcny2016federated} to solve the global problem in \eqref{gl-opt-prob}. In the $m$-th communication round, in parallel each MTD computes the gradient of the local total loss function with respect to the global model parameters, $\nabla\mathcal{L}_k(\mathbf{w}^{(m)}_\textup{g})$, and sends it to the AP. The AP collects all the local gradients and computes the average as
\begin{equation}\label{gl-avg-grad}
 \nabla\mathcal{G}(\mathbf{w}^{(m)}_\textup{g}) = \frac{1}{K} \sum_{k=1}^{K} \nabla \mathcal{L}_k(\mathbf{w}^{(m)}_\textup{g}),
\end{equation}
and distributes the global gradient among all MTDs. Then, each MTD using its dataset $\mathcal{D}_k$ solves the following local loss minimization problem using gradient method:
\begin{equation}\label{loc-opt-prob}
\begin{aligned}
& \underset{ \mathbf{h}^{(m)}_k \in \mathbb{R}^v} {\textup{minimize}}
& & \mathcal{F}_k(\mathbf{w}^{(m)}_\textup{g}, \mathbf{h}^{(m)}_k) = \mathcal{L}_k(\mathbf{w}^{(m)}_\textup{g} + \mathbf{h}^{(m)}_k) 
& & & - \left(\nabla\mathcal{L}_k(\mathbf{w}^{(m)}_\textup{g}) 
- \xi\nabla\mathcal{G}(\mathbf{w}^{(m)}_\textup{g})\right)^\intercal \mathbf{h}^{(m)}_k, \\
\end{aligned}
\end{equation}
\noindent where $(\cdot)^\intercal$ denote the transpose operation, $\xi>0$ is a constant parameter, and $\mathbf{h}^{(m)}_k$ is the model update of the $k$-th MTD's local model parameters in the $m$-th communication round, i.e., $\mathbf{w}^{(m)}_\textup{g} + \mathbf{h}^{(m)}_k$. After local model computation, each MTD sends both the gradient $\nabla\mathcal{L}_k(\mathbf{w}^{(m)}_\textup{g})$ and the update vector $\mathbf{h}^{(m)}_k$ to the AP. Once all local updates are received, the AP computes the global model as follows:
\begin{equation}\label{gl-update-model}
 \mathbf{w}^{(m+1)}_\textup{g} = \mathbf{w}^{(m)}_\textup{g} + \frac{1}{K} \sum_{k=1}^{K} \mathbf{h}^{(m)}_k,
\end{equation}
\noindent and broadcasts the global model towards all MTDs. After sufficient number of communication rounds, the objective function converges to a global optimal \cite{7-SchPol}. Nevertheless, there are many challenges associated with FL's convergence rate in a wireless network scenario with resource-constraint MTDs. For example, the energy-constraint puts a limit on the amount of data that can be sensed and trained. Secondly, the random nature of wireless medium makes the transmission of local model parameters challenging.

\vspace{-0.1cm}
\subsection{Differential Privacy}\label{sec-DP}
 
To protect the sensitive data of the MTDs from the likes of inference or membership attacks, the AP and MTDs preserve the privacy of their computed model parameters by employing DP framework \cite{SnS-Dwork2014b}. As such, a differentially private algorithm allows releasing some statistic or data with provable privacy protection guarantee against arbitrary adversaries. This is achieved by adding suitable level of randomness such that the absence of one individual/entity from a database will not substantially change the distribution of the output of the algorithm \cite{SnS-Dwork2014b, SnS-Nissim}. Before we define DP, let us define the Hamming distance which will be used to measure the similarity of two databases.  

\begin{definition}\label{def-hamming}
Hamming distance between two databases $\mathcal{D}_k \in \mathcal{R}^{d_k}$ and $\mathcal{D}_k' \in \mathcal{R}^{d_k}$ is $\hbar(\mathcal{D}_k, \mathcal{D}_k') = |\{c~|~\mathcal{D}_{k,c} \neq \mathcal{D}_{k,c}'\}|$, which equals to the number of entries where $\mathcal{D}_k$ and $\mathcal{D}_k'$ differ.
\end{definition}

\begin{definition}\label{def-DP}
Let $\mathcal{D}_k = ( \mathcal{D}_{k,1}, \mathcal{D}_{k,2}, \cdot\cdot\cdot , \mathcal{D}_{k,d_k}) \in \mathcal{R}$. For $\epsilon \in (0, 1)$ and $\delta > 0$, a mechanism $\mathcal{M}(\mathcal{D}_k): \mathcal{R}^{d_k} \rightarrow \mathcal{R}$, guarantees $(\epsilon, \delta)$-Differential Privacy if for all sets $\mathcal{S}$, and all parallel databases $\mathcal{D}_k$ and $\mathcal{D}_k'$ which differ by one entry, i.e., $\hbar(\mathcal{D}_k, \mathcal{D}_k') = 1$, we have \cite{dwork2008differential}
\begin{equation}
    p\{\mathcal{M}(\mathcal{D}_k') \in \mathcal{S}~|~\mathcal{D}_k'\} \leq \exp(\epsilon) p\{\mathcal{M}(\mathcal{D}_k) \in \mathcal{S}~|~\mathcal{D}_k\} + \delta,
\end{equation}
\noindent where $p\{\cdot\}$ denote the probability. 
\end{definition}

In Definition \ref{def-DP}, the $\epsilon$ is the privacy budget, i.e., a small value for $\epsilon$ implies more privacy and vice versa. $\delta$ is a very small probability of leaking more information than $\epsilon$. The mechanism $\mathcal{M}$ is the strategy of adding randomness in the algorithm's output in order to achieve privacy. We employ Gaussian mechanism \cite{SnS-Dwork2014b}, to achieve $(\epsilon, \delta)$-DP, which draws a random noise vector from the Gaussian distribution with zero mean and scaled variance such that the privacy of the output statistic vector (model parameters) is preserved. The variance of the noise distribution depends on the ``sensitivity" of the output statistic vector which captures how much a single individual's data can change the output vector. Consider an algorithm applies a function $f: \mathcal{R}^{|\mathcal{D}_k|} \rightarrow \mathcal{R}$ on database $\mathcal{D}_k$ then the Gaussian mechanism on $f$ can be given as
\begin{equation}\label{G-mech}
    \mathcal{M}(\mathcal{D}_k)= f(\mathcal{D}_k) + \mathcal{N}(0,S_f^2\sigma^2_k),
\end{equation}
\noindent where $\mathcal{N}(0,S_f^2\sigma^2_k)$ is Gaussian distribution with zero mean and variance $S_f^2\sigma^2_k$ and $S_f$ is the sensitivity of function $f$. We consider $L_2$-sensitivity and define it as follows:

\begin{definition}\label{def-sense}
The $L_2$-sensitivity of any function $f$ applied on databases $\mathcal{D}_k$ and $\mathcal{D}_k'$ which differ by one entry is
\begin{equation}\label{sensitivity}
    S_f  = \underset{\hbar(\mathcal{D}_k, \mathcal{D}_k') = 1} {\textup{sup}} ||f(\mathcal{D}_k)-f(\mathcal{D}_k')||_2.
\end{equation}
\end{definition}

\begin{proposition}\label{propo-privacy}
Gaussian mechanism $\mathcal{M}$ on function $f$ with sensitivity $S_f$ applied to database $\mathcal{D}_k$ achieves $(\epsilon, \delta)$-DP if 
\begin{equation}\label{sigma}
    \sigma_k   \geq \frac{1}{\epsilon} \sqrt{2\log \Big(\frac{1.25}{\delta}\Big) },
\end{equation}

\noindent where $\epsilon \in (0,1)$ and $\delta>0$.\label{section-DP}
\end{proposition}
\begin{IEEEproof}
The proof is given in \cite{SnS-Dwork2014b}.
\end{IEEEproof}

\vspace{-0.1cm}
\subsection{Differentially Private Federated Learning Algorithm}

The Algorithm 1 specifies the set of operations to be executed at the AP and the MTDs to achieve privacy-preserving FL. In each global communication round, the MTDs follow gradient method to solve local loss minimization problems and compute local model updates. In order to protect local data privacy, the MTDs employ DP Gaussian mechanism as specified in Section~\ref{sec-DP}. Accordingly, each MTD adds a noise vector $\mathbf{n}_k$ drawn from a Gaussian distribution $\mathcal{N}(0,S^2_{f_k}\sigma^2_k)$, where $S_{f_k}$ is the sensitivity of the local update vector $\mathbf{h}^{(m)}_k$ which is given by \eqref{sensitivity}. The objective is to conceal whether a given training sample was used or not by the $k$-th MTD to compute $\mathbf{h}^{(m)}_k$ in the $m$-th communication round. Similarly, the AP protects the global model by adding a noise vector $\mathbf{n}_\textup{g}$ drawn from a Gaussian distribution $\mathcal{N}(0,S^2_{f_\textup{g}}\sigma^2_\textup{g})$, where $S_{f_\textup{g}}$ is the sensitivity of the global model vector $\mathbf{w}^{(m)}_\textup{g}$, and $\sigma^2_\textup{g}$ given by \eqref{sigma}. At the AP, the objective is to conceal whether a given MTD contributed to the global model computation or not. The variance for these noise distributions is controlled through $\sigma^2_k$ and $\sigma^2_\textup{g}$ to achieve $(\epsilon_k, \delta_k)$-DP and $(\epsilon_\textup{g}, \delta_\textup{g})$-DP at the $k$-th MTD and AP, respectively, using Proposition 1.

\begin{algorithm}[t]
\caption{\textit{Differentially Private Federated Learning Algo.}}
\begin{algorithmic}[1]
\STATE \textbf{Inputs:} databases $\{\mathcal{D}_k\}, \forall\, k \in \mathcal{K}$, local problem accuracy threshold = $\Phi$, global problem accuracy threshold $\Psi$.
\STATE \textbf{Initialize:} global communication round number $m=0$, $\mathbf{w}^{(m)}_\text{g}=\mathbf{0}$.
\STATE \textbf{repeat}
\FOR{each MTD $k \in \mathcal{K}$ in parallel}
    \STATE compute $\nabla\mathcal{L}_k(\mathbf{w}^{(m)}_\textup{g})$ and send it to the AP.
\ENDFOR
\STATE AP collects local gradients $\{\nabla\mathcal{L}_k(\mathbf{w}^{(m)}_\textup{g})\}, \forall\, k \in \mathcal{K}$ and computes global average \eqref{gl-avg-grad} and sends it to all MTDs.
\FOR{each MTD $k \in \mathcal{K}$ in parallel}
    \STATE \textbf{Initialize:} local iteration number $j=0$, $\mathbf{h}^{(m),(j)}_k = \mathbf{0} $. 
    \STATE \textbf{repeat}
    \STATE Find the solution $\mathbf{h}^{(m),(j)}_k$ for the local problem in \eqref{loc-opt-prob}.
    \STATE Increment the local iteration number: $j \leftarrow j+1$.
    \STATE \textbf{until} local problem in \eqref{loc-opt-prob} is solved with accuracy $\Phi$.
    \STATE Draw a noise vector $\mathbf{n}_k \in \mathbb{R}^v$ from $\mathcal{N}(0,S^2_{f_k}\sigma^2_k)$ and send noisy solution to the AP, 
    \begin{equation*}
        \mathbf{h}^{(m)}_k \leftarrow \mathbf{h}^{(m),(j)}_k + \mathbf{n}_k.
    \end{equation*}
\ENDFOR
\STATE AP collects all the local model updates and computes the global model \eqref{gl-update-model}.
\STATE AP draws a noise vector $\mathbf{n}_\textup{g} \in \mathbb{R}^v$ from $\mathcal{N}(0,S^2_{f_\textup{g}}\sigma^2_\textup{g})$ and sends noisy global model to MTDs,
    \begin{equation*}
        \mathbf{w}^{(m)}_\textup{g} \leftarrow \mathbf{w}^{(m)}_\textup{g} + \mathbf{n}_\textup{g}.
    \end{equation*}
\STATE Increment the communication round number: $m \leftarrow m+1$.
\STATE \textbf{until} global problem in \eqref{gl-opt-prob} is solved with accuracy $\Psi$.
\end{algorithmic}
\end{algorithm}


Each global communication round in Algorithm 1 is $(\epsilon, \delta)$-DP. It means that the privacy loss of each MTD will be no more than $\epsilon$ with probability greater than or equal to $1-\delta$. Using the composability property of the DP mechanism, we can compute an upper-bound on the total privacy loss for the whole training process by accumulating the privacy loss for each time the training data is accessed, i.e., $\epsilon$. Accordingly, the  Algorithm~1 will be $(M\epsilon, M\delta)$-DP, where $M$ is the total number of global communication rounds. 

For a setting when the same training data is accessed for multiple iterations, recent studies \cite{SnS-Dwork2014b, Dwork2010, Dwork2016, Kasiviswanathan2011, abadi} have analysed the total privacy loss for multiple iterations and proposed much tighter bounds on the privacy loss. For example, based on the strong composition theorem proposed in \cite{Dwork2010} and \cite{Dwork2016} the Algorithm~1 will be $(\epsilon\sqrt{M\log(1/\delta)}, M\delta)$-DP. Moreover, if the training data is sampled from the dataset with probability $q$, then from the privacy amplification theorem proposed in \cite{Kasiviswanathan2011} Algorithm~1 will be $(\mathcal{O}(q\epsilon),\mathcal{O}( q\delta))$-DP. In \cite{abadi}, the proposed moments accountant method achieves a much tighter bound, i.e., $(\mathcal{O}(q\epsilon\sqrt{M}),\mathcal{O}( q\delta))$-DP, which saves a $\sqrt{\log(1/\delta)}$ factor in the $\epsilon$ part and a $Mq$ factor in the $\delta$ part.

\vspace{-0.25cm}
\subsection{Local Model Computation}

\textbf{Computation}:
The MTDs employ gradient method to solve the local loss minimization problem in \eqref{loc-opt-prob}. In a communication round, each MTD performs multiple local iterations following the gradient method. Accordingly, the local update in the $(j~+~1)$-th iteration for the $k$-th MTD given the global model $\mathbf{w}^{(m)}_\textup{g}$ is computed as:
\begin{equation}\label{loc-update-model}
 \mathbf{h}^{(m),(j+1)}_k = \mathbf{h}^{(m),(j)}_k - \eta \nabla \mathcal{F}_k (\mathbf{w}^{(m)}_\textup{g}, \mathbf{h}^{(m),(j)}_k),
\end{equation}
\noindent where $\eta>0$ is the step size. We initialize the local update in each communication round as $\mathbf{h}^{(m),(0)}_k =0,~\forall\,m$. 
Let the $k$-th MTD performs $j$ number of local iterations on its training data of size $d_k$. The computation time to compute local update can be given as:
\begin{equation}\label{loc-cp-time}
 T_{\text{cp},k} = j d_k \tau_k,
\end{equation}
\noindent where $\tau_k$ is given as 
\begin{equation}\label{tau-k}
 \tau_k =  \underbrace{\frac{ \textrm{instructions}}{\textrm{program}}}_{(i)} \times \underbrace{\frac{\textrm{clocks}}{\textrm{instruction}}}_{(ii)} \times \underbrace{\frac{\textrm{seconds}}{\textrm{clock}}}_{(iii)} \times \underbrace{\frac{1}{\textrm{reg}}}_{(iv)},
\end{equation}
\noindent The explanation for the terms in \eqref{tau-k} is as follows:
\begin{description}
  \item [$(i)$] The number of instructions executed to process one data sample of given size. 
  \item [$(ii)$] Since, a typical sensor MTD can execute most of the instructions in 1 clock cycle, we assume the same.
  \item [$(iii)$] Seconds per clock is the clock speed which is equal to the reciprocal of micro- controller's operational frequency (typically in mega or giga Hz).
  \item [$(iv)$] $\textrm{reg}$ is register size of the micr-controller and its value is typically 32-bits or more.
\end{description}

Let $P_{\textup{cp},k}$ denote the power consumed by the $k$-th MTD during data processing. Its predefined value depends on the MTD's hardware. We assume this value is known and constant.

\vspace{-0.1cm}
\subsection{Wireless Transmission}

\textbf{Channel model:}
The AP and all the MTDs are equipped with an omnidirectional antenna. The AP allocates orthogonal radio access channel resources to MTDs for uplink transmission in a given time slot. We assume narrow-band quasi-static propagation channel between each MTD and the AP. Each channel is affected by a large-scale path loss, with path loss exponent~$\alpha$, and a small-scale Rayleigh fading, with channel coefficient $h_k$ for the $k$-th MTD. The channel gain distribution has the scale parameter $\varsigma$. We assume the channel remains unchanged over a single transmission block. In practical networks typically operating at (typically $<$~6~GHz), the coherence time is of the order of 100 ms and the channel estimation is highly accurate \cite{tse-book}. Therefore, we assume the AP estimated the instantaneous channel state for each channel perfectly, which is inline with the relevant prior works \cite{rui2014, george2017, sheeraz-2018}. At the receiver antenna, the noise is assumed to be additive white Gaussian noise with zero mean and variance $\sigma^2_\text{awgn}$. Let $N_0$ denote the noise spectral density. 

\textbf{Transmission:}
Once the local model is computed, each MTD transmits its local model update to the AP using the orthogonal resource blocks, i.e., multiple MTDs can transmit within a given time slot. Recall that each MTD needs to perform both computation and transmission within time block $T$. Let $\mathcal{V}_v$ denote the size of the local model update in bits, which is the same in each communication round for each MTD, i.e., $v$ is fixed for $\mathbf{h}^{(m)}_k \in \mathbb{R}^v,~\forall\,k,m$. Accordingly, the transmission time for the $k$-th MTD, $T_{\textup{tx},k}$, is controlled through link transmission rate, $R_k$, i.e., 
\begin{equation}\label{t-comm}
  T_{\textup{tx},k} =  \frac{\mathcal{V}_v} {R_k}.
\end{equation}
\noindent The transmission rate, $R_k$, is given as
\begin{equation}\label{r}
R_k =  B_k \log_2 \Big(  1 + \frac{\gamma_k}{\Gamma} \Big) ,
\end{equation}
\noindent where $B_k$ is the $k$-th MTD's allocated bandwidth, $\gamma_k$ is the received signal-to-noise ratio (SNR) for the $k$-th MTD, and $\Gamma$ characterizes the gap between the achievable rate and the channel capacity due to the use of practical modulation and coding schemes \cite{wu2016, rui2014}. The received SNR for the $k$-th MTD, $\gamma_k$, is defined as \cite{goldsmith2005wireless}
\begin{equation}\label{snr}
 \gamma_i = \kappa \frac {P_k |h_k|^2}{\sigma^2_\text{awgn} r_k^\alpha},
\end{equation}
\noindent where  $P_k$ is the transmit power for the $k$-th MTD, $\kappa~=~\big(\frac{\text{c}}{4\pi f_c}\big)^2$ is the path loss factor, $\text{c}$ is the speed of light, $f_c$ is the center frequency, and $r_k$ is the distance between MTD and the AP.
To compute the data transmission power cost $P_{\text{tx},k}$ for the $k$-th MTD, we adopt a practical model of \cite{wu2017}. $P_{\text{tx},k}$ is composed of three components: (i)~the transmit power $P_k$, (ii)~amplifier power $P_{\text{amp},k}$, and (iii)~the static communication module circuitry power $P_{\text{cir},k}$, i.e., $P_{\text{tx},k} = P_k + P_{\text{amp},k} + P_{\text{cir},k}$. $P_{\text{cir},k}$ accounts for the operation of the frequency synthesizer, mixer, transmit filter, antenna circuits, digital-to-analog converter, etc. We consider $P_\text{amp} = (1/\rho-1) P_k$, \cite{wu2017}, where $\rho \in (0,1]$ is the drain efficiency of the power amplifier. Thereby, $P_{\text{tx},k}$ for the $k$-th MTD can be given as
\begin{equation}\label{p-tx}
  P_{\text{tx},k} =  \frac{1}{\rho}P_k + P_{\text{cir},k}.
\end{equation}

\section{Proportionally-Fair Differentially-Private Federated Learning}

In the proposed system, the objective of the AP is to learn a statistical model over the datasets of all participating MTDs without acquiring these datasets. Whereas, the objective of each MTD is to learn a better statistical model trained over multiple disjoint datasets of different MTDs as compared to its local model trained over much smaller dataset. Hence, the global model is more accurate than the individual local models. The improvement achieved in statistical model learning at the MTDs after each communication round is referred to as the learning gain. 

The learning gain is achieved by participation in FL mechanism, i.e., performing the local model training and transmission, which poses significant energy cost on resource constrained MTDs. The quality of the local model and the associated cost of computation and transmission will be different for different MTDs, because of the device heterogeneity in the sensing data, wireless channel, and availability of the computation and other resources. In this regard, we quantify the utility of $k$-th MTD in $m$-th communication round, $\mathcal{U}^{(m)}_k$, by the degree of the learning gain and the accumulative energy cost associated with local model computation and transmission, i.e.,
\begin{equation}
 \mathcal{U}^{(m)}_k \propto \{\text{Learning gain}\}^{(m)} - \{\text{Cost}_\text{cp+tx}\}^{(m)}.
\end{equation}

In a vanilla FL setting, each participating MTD receives the same global model update irrespective of the quality of its shared local model and the associated cost. Therein, a malicious MTD may intentional try to reduce the cost of computation and/or transmission of the local model, greedily increasing its utility by enjoying the same global model, i.e., level of accuracy, as other legitimate MTDs. This malicious behavior in a FL setting causes utility unfairness among MTDs and its poor quality local model adversely impacts the quality of the global model which in turn increases the convergence time and reduces the utility of the legitimate MTDs. That is, even if legitimate MTDs increase their computation budget to maximize the quality of the local model to help achieve a better global model, their additional investment doesn’t pay-off productivity gains due to the poor local models contributed by the malicious MTDs. Consequently, the legitimate MTDs don’t attain the expected improvement in their utility. The same problem can naturally be caused due to the device heterogeneity. Wherein, the  local model quality and accumulative cost can be impacted by the (time-varying) diversity in the training data, wireless channel, computation/communication resources, etc. Nevertheless, distinguishing the malicious behaviour from the natural channel/resource diversity is rather hard without additional  information.  

Since, in the state-of-the-art FL settings, there are no specific standard of operations devised for participating MTDs, the AP cannot ensure their pertinent operation. To address these challenges in a FL based wireless IoT network with heterogeneous MTDs, we propose to preserve global model using DP. We design policies for the AP and the MTDs, where the AP strive to ensure utility fairness among MTDs and the MTDs try to maximize their utility. In particular, the utility function works as a catalyst and it is used to reveal the optimal computation and transmission policies, such that the learning gain versus the cost is similar for all devices.

\subsection{MTD's Local Training Convergence and Cost}

We make the following assumptions to facilitate the design and analysis of the proposed system.

\textbf{Assumption 1:} The gradient $\nabla\mathcal{F}_k(\mathbf{w}^{(m)}_\textup{g},\mathbf{h}^{(m),(j)}_k)$ of function $\mathcal{F}_k(\mathbf{w}^{(m)}_\textup{g},\mathbf{h}^{(m),(j)}_k)$, for any $(\mathbf{w}^{(m)}_\textup{g},\mathbf{h}^{(m),(j)}_k)$ in its domain, is $L$-Lipschitz continuous and it holds that
\begin{equation}\label{assume-1}
    ||\nabla\mathcal{F}_k(\mathbf{w}^{(m)}_\textup{g},\mathbf{h}^{(m),(j+1)}_k) -
    \nabla\mathcal{F}_k(\mathbf{w}^{(m)}_\textup{g},\mathbf{h}^{(m),(j)}_k)|| 
    \leq L ||\mathbf{h}^{(m),(j+1)}_k - \mathbf{h}^{(m),(j)}_k||,
\end{equation}
\noindent where $L>0$ is a Lipschitz constant for the function $\nabla\mathcal{F}_k(\cdot)$.

\textbf{Assumption 2:} The function $\mathcal{F}_k(\mathbf{w}^{(m)}_\textup{g},\mathbf{h}^{(m),(j)}_k)$, for any $(\mathbf{w}^{(m)}_\textup{g},\mathbf{h}^{(m),(j)}_k)$ in its domain, is strongly convex with parameter $\mu>0$ and it holds that
\begin{multline}\label{assume-2}
    \mathcal{F}_k(\mathbf{w}^{(m)}_\textup{g},\mathbf{h}^{(m),(j+1)}_k) 
    \geq
    \mathcal{F}_k(\mathbf{w}^{(m)}_\textup{g},\mathbf{h}^{(m),(j)}_k) 
    + \left( \mathbf{h}^{(m),(j+1)}_k - \mathbf{h}^{(m),(j)}_k \right)^\intercal 
    \nabla \mathcal{F}_k(\mathbf{w}^{(m)}_\textup{g},\mathbf{h}^{(m),(j)}_k) 
    \\
    + \frac{\mu}{2} || \mathbf{h}^{(m),(j+1)}_k - \mathbf{h}^{(m),(j)}_k||^2.
\end{multline}

\textbf{Assumption 3:} The function $\mathcal{F}_k(\mathbf{w}^{(m)}_\textup{g},\mathbf{h}^{(m),(j)}_k)$, for any $(\mathbf{w}^{(m)}_\textup{g},\mathbf{h}^{(m),(j)}_k)$ in its domain, is twice-continuously differentiable. Thereby, \eqref{assume-1} and \eqref{assume-2} means that the eigenvalues of the Hessian are uniformly bounded above and below \cite{nesterov2013introductory}, this can be given as
\begin{equation}\label{assume-3}
    \mu \mathbf{I} 
    \preceq
    \nabla^2 \mathcal{F}_k(\mathbf{w}^{(m)}_\textup{g},\mathbf{h}^{(m),(j)}_k) 
    \preceq 
    L \mathbf{I},
\end{equation}
\noindent where $\mathbf{I}$ is the identity matrix and the ratio $L/\mu \geq 1$ is the condition number of function $\mathcal{F}_k(\cdot)$.

All of the above assumptions can be satisfied by various convex loss functions, e.g., the cross-entropy, logistic regression, and mean squared error \cite{32-joint}. These loss functions are usually preferred for the performance analysis of identification, prediction, and classification algorithms.

Following Algorithm~1, in each communication round, all MTDs perform multiple local iterations to solve the local problem in \eqref{loc-opt-prob} with an accuracy of $\Phi$. 

\begin{definition}\label{def-accuracy}
For the $k$-th MTD, in the $m$-th communication round, we define the accuracy $\Phi \in (0,1)$ of the solution $\mathbf{h}^{(m),(j)}_k$ to the local problem in \eqref{loc-opt-prob} after $j$ iterations as
\begin{equation}\label{loc-accuracy}
    \Phi \geq \frac{\mathcal{F}_k(\mathbf{w}^{(m)}_\textup{g},\mathbf{h}^{(m),(j)}_k)-\mathcal{F}_k(\mathbf{w}^{(m)}_\textup{g},\mathbf{h}^{*(m)}_k)  } {\mathcal{F}_k(\mathbf{w}^{(m)}_\textup{g},\mathbf{0})-\mathcal{F}_k(\mathbf{w}^{(m)}_\textup{g},\mathbf{h}^{*(m)}_k)},
\end{equation}
\noindent where $\mathbf{h}^{*(m)}_k$ is the optimal solution of the problem in \eqref{loc-opt-prob}.  
\end{definition}

Based on the assumptions in \eqref{assume-1}, \eqref{assume-2} and \eqref{assume-3}, we obtain a lower bound on the number of iterations, $j_\textup{min}$, required to solve local problem in \eqref{loc-opt-prob} with an accuracy $\Phi$.

\begin{theorem}\label{theorem-iteration}
If the objective function $\mathcal{F}_k(\cdot)$ in \eqref{loc-opt-prob} is twice-continuously differentiable $\mu$-strongly convex and its gradient $\nabla \mathcal{F}_k(\cdot)$ is $L$-Lipschitz continuous, then $k$-th MTD employing the gradient method with a step size $\eta > \frac{L}{2}$ needs to perform 
\begin{equation}\label{iter-bound}
  j_{\textup{min},k} \geq \frac{ \log(\Phi) } 
            { \log \big( \frac{\eta^2 L^2}{2} - \eta L  + 1 \big) }
\end{equation}
\noindent number of iterations to solve the local problem in \eqref{loc-opt-prob} with an accuracy $\Phi$.
\end{theorem}
\begin{IEEEproof}
The proof is provided in Appendix~\ref{A}.
\end{IEEEproof}

\vspace{0.2cm}
Finally, for the $k$-th MTD the energy cost of the local model training and transmission in the $m$-th communication round, denoted by $E^{(m)}_{\text{cp+tx},k}$, can be given as follows:
\begin{equation}\label{e-c-tmp}
  E^{(m)}_{\text{cp+tx},k} = T^{(m)}_{\text{cp},k} P^{(m)}_{\text{cp},k} + T^{(m)}_{\text{tx},k} P^{(m)}_{\text{tx},k}.
\end{equation}
From \eqref{loc-cp-time}, \eqref{t-comm}, and \eqref{p-tx}, we can rewrite \eqref{e-c-tmp} as
\begin{equation}\label{energy-cost}
  E^{(m)}_{\text{cp+tx},k} = j_k d_k \tau_k P_{\text{cp},k} + 
  \frac{\mathcal{V}_v \big(\frac{1}{\rho}P^{(m)}_k + P_{\text{cir},k} \big)} 
  {B_k \log_2 \Big(  1 + \kappa \frac {P^{(m)}_k |h^{(m)}_k|^2}{\sigma^2_\text{awgn} r_k^\alpha\Gamma}\Big)},
\end{equation}
\noindent where $x^{(m)}$, for any variable $x$, indicates that $x$ changes from one round to the next.

\subsection{Utility Fairness Policy for AP}

Based on Theorem 1, we can specifically state that in the $m$-th communication round the $k$-th MTD performs $j$ iterations and solves the problem in \eqref{loc-opt-prob} with accuracy $\Phi$ using its local dataset $\mathcal{D}_k$. The solution $\mathbf{h}_k^{(m),(j)}$, i.e., local model update, is then sent to the AP. After collecting local updates from all participating MTDs, the AP computes the global model $\mathbf{w}^{(m)}_\textup{g}$ update as in \eqref{gl-update-model}, we refer to it as the true global model update. 

As discussed earlier, the quality of the local model update can be significantly different for different MTDs due to various factors. In this regard, if the same global model is shared with all participating MTD, actually different MTDs spend different costs on local model computation and transmission, which would give rise to utility unfairness. In this setting, we propose to control the quality of the global model shared with different MTDs, in each communication round, based on their contribution towards the global model computation. In that the true global model will be treated as a precious commodity and will be shared in a differentially private manner among different MTDs. Accordingly, the AP will add noise in the global model before sharing it with a given MTD in proportion to the deviation of its local model from the global model. In the proposed methodology, the AP relies only on the local model quality to decide the level of noise to be added in each communication round. It is because in practical settings the computation and transmission cost of MTDs cannot not be quantified effectively at the AP.

As per Algorithm~1, a specific level of noise needs to be added to the true global model to achieve $(\epsilon_\text{g},\delta_\text{g})$-DP by drawing noise vector from $\mathcal{N}(0,S^2_{f_\textup{g}}\sigma^2_\textup{g})$. In addition to that, in each communication round, the AP tries to curtail the information about the true global model by proportionally enlarging the variance of the distribution the noise vector is drawn from for individual MTDs. The change in variance is proportional to the quality of the individual MTD's local model update in the given communication round. Recall that a smaller $\epsilon$ means more privacy, i.e., high noise. Accordingly, in the $m$-th communication round the noise vector for the $k$-th MTD is drawn from distribution $\mathcal{N}(0, S^{2,(m)}_{f_\textup{g}}\widehat{\sigma}^{2,(m)}_{\textup{g},k})$, where 
\begin{equation}
    \widehat{\sigma}^{(m)}_{\textup{g},k} \geq \frac{1}{\epsilon_\textup{g} ( 1 - \mathcal{E}^{(m)}_k \theta) } \sqrt{2\log \Big(\frac{1.25}{\delta_\textup{g}}}\Big),
\end{equation}
\noindent and $\mathcal{E}^{(m)}_k \in [0,1]$ captures how different the local model is from the true global model, and $\theta \in [0,1]$ calibrates the impact of $\mathcal{E}^{(m)}_k$ on $\epsilon_\textup{g}$. We refer to $\mathcal{E}^{(m)}_k$ as the deviation factor and employ the Cosine Similarity measure to quantify it as follows:
\begin{equation}\label{dev-factor}
    \mathcal{E}^{(m)}_k = 1 - 
    \frac{ \textup{sim}\big(\mathbf{w}^{(m)}_\textup{g},\mathbf{h}^{(m)}_k\big)}
    {\underset{ \forall \, k \in \mathcal{K}}  {\max} \big\{\text{sim}\big(\mathbf{w}^{(m)}_\textup{g},~\mathbf{h}^{(m)}_k \big) \big\} },
\end{equation}
\noindent where $\textup{sim}(\cdot)$ is the Cosine Similarity operation and $\max\{\cdot\}$ is the max operation. The deviation factor defined in \eqref{dev-factor} ensures proportional utility fairness by increasing the noise distribution variance in a relative manner. Therein, the MTD with the best local model is not penalised with enlarged noise other than the required privacy level to achieve $(\epsilon_\text{g},\delta_\text{g})$-DP. The rest of the MTDs are proportionally penalised as per the quality of their individual local model updates. Accordingly, the global model update sent to the $k$-th MTD in the $m$-th communication round can be given as
\begin{equation}\label{gl-model-k}
    \mathbf{w}^{(m)}_{\textup{g},k} = \mathbf{w}^{(m)}_\textup{g} + \mathbf{n}^{(m)}_k,
\end{equation}
\noindent where $\mathbf{w}^{(m)}_\textup{g}$ is the true global model, $\mathbf{n}^{(m)}_k$ is the noise vector drawn from $\mathcal{N}(0, S^{2,(m)}_{f_\textup{g}}\widehat{\sigma}^{2,(m)}_{\textup{g},k})$. The AP also sends the deviation factor $\mathcal{E}^{(m)}_k$ to imply the contribution of the $k$-th MTD's local model, relative to other MTDs, towards the global model update. Large deviation factor implies lesser contribution. This indicates to MTDs the severity of additive noise in the global model shared by the AP.

\subsection{Utility Maximization Policy for MTDs}

The objective of each MTD is to allocate the optimal budget, for the local model computation and transmission, which will maximize its utility under given system constraints. The utility of a MTD decreases when the AP adds more noise to its global model update, which is directly proportional to the MTD's deviation factor. Hence, the learning gain is proportional to the deviation factor, i.e., local model's quality. The deviation factor increases by solving the local problem with lesser accuracy, or by using only a fraction of the local dataset for training, running less number of iterations. Overall, the deviation factor is directly proportional to the computation cost which is controlled by each MTD to maximize its utility. 

In this regard, each MTD performs the following operations after receiving its global model and the deviation factor from the AP. In expectation of receiving a less noisy global model in the next round, each MTD uses an energy cost to deviation factor model to map the decrease in the deviation factor when the budget for the computation cost is increased. We found this model using data fitting analysis for a FL system where MTDs controlled the computation cost using various ways mentioned above. In this model, the deviation factor expected by the $k$-th MTD in the $m$-th communication round for a given computation cost is given as
\begin{equation}
\mathcal{E}^{(m)}_{\textup{mod}, k} = \beta^{(m)}_{1,k} \exp \Big( - \frac{1}{\beta^{(m)}_{2,k}}
E^{(m)}_{\text{cp},k} \Big),
\end{equation}
\noindent where $E^{(m)}_{\text{cp},k}=T^{(m)}_{\text{cp},k} P^{(m)}_{\text{cp},k}$ is the computation energy cost and $\beta^{(m)}_{1,k}, \beta^{(m)}_{2,k} > 0$ are model parameters. Using this model, the $k$-th MTD estimates the values of $\beta^{(m)}_{1,k}$ and $\beta^{(m)}_{2,k}$ observed in the given communication round as follows:
\begin{equation}
 \hat{\beta}^{(m)}_{1,k}, \hat{\beta}^{(m)}_{2,k} \leftarrow \underset{ \beta^{(m)}_{1,k}, \beta^{(m)}_{2,k}} {\textup{argmin}}
|\mathcal{E}^{(m)}_k - \mathcal{E}^{(m)}_{\textup{mod}, k} |^2,
\end{equation}
\noindent where $\mathcal{E}^{(m)}_k$ is the actual deviation factor observed in the $m$-th communication round by the $k$-th MTD. Note the causality that is $\hat{\beta}^{(m)}_{1,k}$ and $\hat{\beta}^{(m)}_{2,k}$ are revealed at the end of a given communication round. These values of $\hat{\beta}^{(m)}_{1,k}$ and $\hat{\beta}^{(m)}_{2,k}$ are then used by the MTD in the next $(m+1)$-th communication round to maximize its expected-utility. This strategy of MTD closely resembles the risk-aversion in expected-utility theory \cite{exp-utility}, wherein the utility function is modelled as concave in cost. Accordingly, we model the utility function of the $k$-MTD as the following concave function:
\begin{equation}\label{MTD-utility}
\mathcal{U}_k = - \mathcal{E}_{\textup{mod}, k} + \beta_{1,k} - E_{\textup{cp+tx}, k} (E_{\textup{cp+tx}, k} - \varrho),
\end{equation}
\noindent where parameter $\varrho > 0$ captures the relationship between the utility and MTD's energy cost. In \eqref{MTD-utility}, the first two terms jointly represent the relative quality of the local model and the other term represents the impact of the total energy cost.

Finally, the MTD follows the optimal computation and transmission policies which will yield the maximum expected-utility in the next $(m+1)$-th communication round as per the given estimates of $\beta_{1,k}$ and $\beta_{2,k}$ from previous round, i.e., $\hat{\beta}^{(m)}_{1,k}$ and $\hat{\beta}^{(m)}_{2,k}$. The computation policy determines the optimal number of iterations to perform to solve the local problem with an optimal level of accuracy which will maximize utility. Whereas, the transmission policy determines the optimal link transmission rate to be controlled through the transmit power. Both of these policies must satisfy all system constraints. This utility maximization problem for the $k$-MTD can be formulated as follows
\begin{subequations}\label{utility-max-opt-prob-tmp}
\begin{alignat}{2}
& \hspace{-1.0cm}\underset{ \substack{\Phi^{(m+1)}_k,~j^{(m+1)}_k, \\ P^{(m+1)}_k,~R^{(m+1)}_k} }   {\textup{maximize}}
& &  \mathcal{U}_k\big(\hat{\beta}^{(m)}_{1,k}, \hat{\beta}^{(m)}_{2,k}, j^{\S,(m+1)}_k, P^{(m+1)}_k \big)         \label{utility-max-opt-prob-tmp-a}                  \\
& \hspace{-0.6cm}\textup{subject to}
& &    T^{(m+1)}_{\textup{cp}, k} + T^{(m+1)}_{\textup{tx}, k} \leq T, \label{utility-max-opt-prob-tmp-b} \\
& & &  P_{\textup{min}, k} \leq P^{(m+1)}_k \leq P_{\textup{max}, k}, \label{utility-max-opt-prob-tmp-d}\\
& & &  j_{\textup{min}, k} \leq j^{(m+1)}_k \leq j_{\textup{max}, k}, \label{utility-max-opt-prob-tmp-f}\\
& & &  0 \leq \Phi^{(m+1)}_k \leq 1, \label{utility-max-opt-prob-tmp-e}\\
& & &  0 \leq R^{(m+1)}_k. \label{utility-max-opt-prob-tmp-g}%
\end{alignat}
\end{subequations}
\noindent where the constraint \eqref{utility-max-opt-prob-tmp-b} states that accumulative local model computation and transmission time should not exceed the delay bound. The remaining constraints reflect practical range of values for the design variables, where $P_{\textup{min}, k}$ is the transmit power of the $k$-th MTD required to perform $j_{\textup{min}, k}$ iterations for the given channel realization and delay bound, i.e., 
\begin{equation}\label{P-min}
P_{\textup{min}, k} =  \frac {\sigma^2_\textup{awgn} r^\alpha_k \Gamma } {\kappa |h_k|^2}  \bigg( \exp \Big( \frac{\mathcal{V}_v \log(2)} 
  {B_k (T - j_{\textup{min}, k} d_k \tau)} \Big) -  1 \bigg),
\end{equation}
\noindent and $j_{\textup{max}, k}$ is the upper bound on the number of iterations the $k$-th MTD can perform for the given channel realization and delay bound, which can be given as
\begin{equation}\label{j-max}
j_{\textup{max}, k} = \frac{1}{d_k \tau_k} \Big( T  {-}  \frac{\mathcal{V}_v \log(2) } 
  {B_k \log \big(  1 {+} \kappa \frac {P_{\textup{max}, k} |h_k|^2}{\sigma^2_\text{awgn} r_k^\alpha\Gamma}\big)} \Big).
\end{equation}

\begin{remark}\label{remark-1}
Using Theorem 1, we can compute the bound on the accuracy with which the local problem will be solved after a given number of iterations. Similarly, the link transmission rate can be computed through the transmit power.
\end{remark}

Based on Remark 1, substituting $T^{(m+1)}_{\textup{cp}, k}$, $T^{(m+1)}_{\textup{tx}, k}$, and $P^{(m+1)}_{\textup{tx}, k}$ from \eqref{loc-cp-time}, \eqref{t-comm}, and \eqref{p-tx}, for an arbitrary communication round, the problem \eqref{utility-max-opt-prob-tmp} can equivalently be given as:
\vspace{-0.2cm}
\begin{subequations}\label{utility-max-opt-prob}
\begin{alignat}{2}
& \underset{ j_k,~P_k }  {\textup{maximize}}
& & \hspace{0.3cm} \mathcal{U}_k\big(\hat{\beta}_{1,k}, \hat{\beta}_{2,k}, j_k, P_k \big)                         \label{utility-max-opt-prob-a}  \\
& \textup{subject to}
& & \hspace{0.3cm}   j_k d_k \tau_k {+} 
  \frac{\mathcal{V}_v \log(2) } 
  {B_k \log \big(  1 + \kappa \frac {P_k |h_k|^2}{\sigma^2_\text{awgn} r_k^\alpha\Gamma}\big)} \leq T, \label{utility-max-opt-prob-b}\\
& & & \hspace{0.3cm}  P_{\textup{min}, k} \leq P_k \leq P_{\textup{max}, k}, \label{utility-max-opt-prob-d}\\
& & & \hspace{0.3cm} j_{\textup{min}, k} \leq j_k \leq j_{\textup{max}, k}.\label{utility-max-opt-prob-e}
\end{alignat}
\end{subequations}
It can be shown that the problem in \eqref{utility-max-opt-prob} is a non-convex optimization problem.

\begin{lemma}\label{lemma-convex-transform}
The optimization problem in \eqref{utility-max-opt-prob} can be transformed into an equivalent convex problem. Thus, a globally optimal solution
exists for the problem in \eqref{utility-max-opt-prob}.
\end{lemma}

\begin{IEEEproof}
The proof is provided in Appendix \ref{B}.
\end{IEEEproof} 

From Lemma \ref{lemma-convex-transform}, we have the following equivalent convex problem for \eqref{utility-max-opt-prob}:
\begin{subequations}\label{utility-max-opt-prob-convex}
\begin{alignat}{2}
& \underset{ j_k,~Z_k }  {\textup{minimize}}
& & \hspace{0.3cm} -\mathcal{U}_k\big(\hat{\beta}_{1,k}, \hat{\beta}_{2,k}, j_k, Z_k \big)                         \label{utility-max-opt-prob-a}  \\
& \textup{subject to}
& & \hspace{0.3cm}   j_k d_k \tau_k + \frac{\mathcal{V}_v \log(2)}  {B_k Z_k} \leq T, \label{utility-max-opt-prob-b}\\
& & & \hspace{0.3cm}  Z_{\textup{min},k} \leqslant Z_k \leqslant Z_{\textup{max},k}, \label{utility-max-opt-prob-d}\\
& & & \hspace{0.3cm}  j_{\textup{min}, k} \leq j_k \leq j_{\textup{max}, k}.\label{utility-max-opt-prob-e}
\end{alignat}
\end{subequations}
\noindent where $Z_k = \log \big(  1 + \frac { \kappa P_k |h_k|^2}{\sigma^2_\textup{awgn} r^\alpha_k \Gamma } \big)$, $Z_{x,k} = \log \big(  1 +  \frac {\kappa P_{x,k} |h_k|^2}{\sigma^2_\textup{awgn} r^\alpha_k \Gamma } \big)$, $x \in \{\text{min, max\}}$. \vspace{0.1cm} 

\begin{remark}\label{remark-2}
The solution to the problem in \eqref{utility-max-opt-prob-convex}, i.e., the optimal values of $j_k$ and $Z_k$, yield the optimal solution to the problem in \eqref{utility-max-opt-prob-tmp} which will maximize its objective function.
\end{remark}

Based on Remark \ref{remark-1} and Lemma \ref{lemma-convex-transform}, the solution to the problem in \eqref{utility-max-opt-prob-convex} yields the solution to the problem in \eqref{utility-max-opt-prob-tmp} as given by the following theorem.
\begin{theorem}
In solving the optimization problem \eqref{utility-max-opt-prob-convex}, the optimal transmission rate, $R^*_k$, is given by
\begin{equation}\label{optimal-R}
R^*_k = B_k \log_2 \Big(  1 {+} \kappa \frac {P^*_k |h_k|^2}{\sigma^2_\textup{awgn} r^\alpha_k \Gamma } \Big)
\end{equation}
\noindent where the optimal transmit power, $P^*_k$, is given by
\begin{equation}\label{optimal-P}
P^*_k = \frac {\sigma^2_\textup{awgn} r^\alpha_k \Gamma } {\kappa |h_k|^2}  \big( \exp(Z^*_k) -  1 \big),
\end{equation}
\noindent where
\begin{equation}\label{optimal-z}
Z^*_k=
    \begin{cases}
      \max \big(Z_{\textup{min}, k},~\hat{Z}_k \big),    
      & \textup{if} \; \hat{Z}_k < Z_{\textup{max}, k}, \\
      Z_{\textup{max}, k},      &   \textup{otherwise,}
    \end{cases}
\end{equation}
\noindent where $\hat{Z}_k$ is given by numerically solving following equality 
\begin{multline}\label{optimal-z-hat}
\bigg( 2 P_{\textup{cp},k} \Big( T - \frac{\mathcal{V}_v \log(2)}  {B_k \hat{Z}_k}  \Big)
+  \frac{2 \mathcal{V}_v b_k} {\hat{Z}_k}  \big( \exp  ( \hat{Z}_k ) + c_k  \big) - \varrho  \bigg)
\\
\Big( \frac {B_k b_k}{P_{\textup{cp},k} \log(2)} \big( {(\hat{Z}_k - 1) \exp( \hat{Z}_k ) - c_k} \big) + 1 \Big)
\\
= \frac{\beta_{1,k}}{\beta_{2,k}} \exp \bigg( \frac{P_{\textup{cp},k}}{\beta_{2,k}}
  \Big(  \frac{\mathcal{V}_v \log(2)}  {B_k \hat{Z}_k} - T \Big) \bigg),
\end{multline}
\noindent where 
\begin{equation*}
b_k = \frac{ \sigma^2_\textup{awgn} r^{\alpha}_k \Gamma \log(2)}{\rho \kappa B_k |h_k|^2},
%
\; c_k = \frac{\rho \kappa |h_k|^2 P_{\textup{cir},k} } {  \sigma^2_\textup{awgn} r^{\alpha}_k \Gamma} - 1,
\end{equation*}
\noindent and the optimal number of iterations, $j^*_k$, to perform is given by 
\begin{equation}\label{optimal-j}
j^*_k = \min \Big(\frac{1}{d_k \tau_k} \big( T {-} \frac{\mathcal{V}_v \log(2)}  {B_k Z^*_k} \big),~j_{\textup{max}, k}\Big),
\end{equation}
\noindent and the optimal accuracy, $\Phi^*_k$, is bounded by
\begin{equation}\label{optimal-phi}
 \Phi^*_k \leq \exp \Big( \big( j^*_k + 1 \big)  \log \big( \frac{\mu \eta^2 L}{2} - \mu \eta  + 1 \big) \Big)
\end{equation}
\end{theorem}
\begin{IEEEproof}
The proof is provided in Appendix~\ref{C}.
\end{IEEEproof}

\section{Simulation Results}\label{sec-results}

This section presents simulation results to illustrate the performance of the proposed scheme. We first present the learning performance of the proposed scheme, and then study the impact of device heterogeneity on the energy expenditure and the trade-off between computation and transmission cost. To model the variable computation cost of local training at the MTDs, we keep the dataset size same at each MTD and the quality of the local model is controlled through the number of iterations, which is inline with prior works \cite{personal, 9253545, 9317806}. 

For simulations we consider the MNIST dataset available for digit recognition task~\cite{MNIST} and a neural network for training with an input layer with 784 units, two hidden layers (the first with 128 units, the second with 64 units) each using ReLu activation, then an output layer with 10 units, and the softmax output. The total number of parameters is 109,375, each represented by one byte, i.e., $\mathcal{V}_v$ = 875 kbits. The batch size is set to 128 for all MTDs. The data is partitioned among MTDs in an i.i.d. fashion. Gradient method is used for optimizing with negative log likelihood loss and a learning rate of 0.01. The sensitivity of the data varies from one communication round to the next and its value fluctuates around 0.01. Unless specified otherwise, the parameters values shown in Table \ref{para-table} are adopted. The joint use of regularization and DP fend off overfitting, which improves the model accuracy for the data it has not seen before. Thus, the training loss is expected not to drop to zero.
Note that the FL performance with DP has been well studied in previous studies and thus it is not the focus of this paper. Our focus is rather to evaluate the relative performance of the heterogeneous wireless MTDs in a differentially-private FL algorithm.

\begin{table*}[]
\centering
\caption{System parameter values.}
\label{para-table}
\begin{tabular}{|l|c|c||l|c|c|} \hline
\textbf{Name}               & \textbf{Symbol} & \textbf{Value} & \textbf{Name}   & \textbf{Symbol}  & \textbf{Value}  \\ \hline
Global privacy budget     & $\epsilon_\text{g}$ & 0.95     & Circuitry power     & $P_{\textup{cir},k}$ & 82.5 mW, $\forall k$ \\
Global leakage probability& $\delta_\text{g}$  & $10^{-5}$ & Power bound         & $P_{\textup{max},k}$ & 0 dB, $\forall k$   \\
Number of MTDs            & $K$                 & 10       & Delay bound           & $T$                & 0.75 ms       \\
Pathloss exponent         & $\alpha$            & 4        & Per-bit processing time & $\tau$           & 7.5 ns/b   \\
Modulation power gap      & $\Gamma$            & 9.8 dB   & Local model size      & $\mathcal{V}_v$    & 875 kbits     \\
Utility-Energy  parameter & $\varrho$           & 0.5      & Bandwidth             & $B$                & 250 KHz     \\
Fading parameter          & $\varsigma$         & 1        &    Min. no. of iterations    & $j_{\textup{min},k}$    & 10, $\forall k$ \\
Center frequency          & $f_c$               & 32 MHz   & Distance AP-MTD       & $r_k$              & \{50-200\} m, $\forall k$ \\
Privacy scale parameter   & $\theta$            & 0.6      & Computation power     & $P_{\textup{cp},k}$& 96 mW, $\forall k$ \\
Amplifier efficiency      & $\rho$              & 0.45     &Noise spectral density & $N_0$              & $-$174 dBm       \\ \hline
%
%
\end{tabular}
\end{table*}

To the best of our knowledge, the recent works \cite{personal, 9253545, 9317806} are the most relevant to our proposed scheme. Although the objectives are different, the system models specifying the underlying FL and DP implementation are similar to our considered system. In this regard, our objective is to identify the utility unfairness issue among MTDs and these models suffice to demonstrate it. Once unfairness issue is divulged, we analyse the performance of the proposed scheme to counter that. When our considered system is applied, the design for the DP based FL models in \cite{personal, 9253545, 9317806} can equivalently be represented by the following benchmark scheme.

\textbf{\textit{Benchmark scheme}}:
For the benchmark scheme, the algorithm aims to achieve a fixed $(\epsilon_\text{g},\delta_\text{g})$-DP for both local and global models sharing. This is the minimum level of DP the proposed scheme already guarantees. In the benchmark scheme, each MTD tries to perform the maximum number of iterations ($\leq j_\textup{max}$) possible and transmit the noisy local model to the AP under the given channel realization, the delay constraint. The AP receives the local models and generate the global model. A noisy version of this global model, which is same for each MTD, providing $(\epsilon_\text{g},\delta_\text{g})$-DP is then sent to all MTDs. The strategy followed by the benchmark scheme implements a generic differentially-private FL algorithm with computation and transmission cost control for a wireless network. This strategy is essentially the same as in the state-of-the-art in \cite{personal, 9253545, 9317806}. The corresponding optimization problem for the benchmark scheme is omitted here for brevity. For fair comparison with the proposed scheme, the dataset of a given MTD and channel realizations in a given round are kept the same for both schemes.

\begin{figure}[t]
\centering
    \begin{tikzpicture}[spy using outlines={ chamfered rectangle, magnification=2.0, width=1.25cm, height=2.0cm,, connect spies}]
    \begin{axis}[
     each nth point=1,
     height=6.5cm, width=8.0cm,
     legend cell align=left,
     legend style={inner xsep=1pt, inner ysep=1pt,at={(0.95,0.85)},anchor=east,font=\small, legend columns=1, draw, fill},
     cycle list name = mycyclelist_no_marks,
     mark repeat={1},
     grid=both,
     label style={font=\small},
     xtick pos=left,
     ytick pos=left,
     xmin=0, xmax=50, xtick={0,5,10,15,20,25,30,35,40,45,50},
     xlabel= Communication rounds,
     ylabel= Avg. total train loss: $\frac{1}{K}\sum\limits_{i=1}^{K} \mathcal{L}^{(m)}_k$,
     ylabel style={yshift=-0.5ex},
     ymin=0.4, ymax=2.4, ytick={0.4,0.6,0.8,1.0,1.2,1.4,1.6,1.8,2.0,2.2,2.4},
     ticklabel style={
        /pgf/number format/fixed,
        /pgf/number format/precision=5
     }
     ]
             \addplot+ [
                error bars/.cd,
                    y explicit,
                    y dir=both,
            ] table [
                x=Round,
                y=NUM,
            ] {data_b.txt};
            
            \addplot+ [
                error bars/.cd,
                    y explicit,
                    y dir=both,
            ] table [
                x=Round,
                y=NUM,
                y error plus expr=\thisrow{CI-H}-\thisrow{NUM},
                y error minus expr=\thisrow{NUM}-\thisrow{CI-L},
            ] {data_p.txt};
    \legend{Benchmark scheme, Proposed scheme}
    \end{axis}
        \end{tikzpicture}
\caption{Average MTD total train loss, $\frac{1}{K}\sum_{i=1}^{K} \mathcal{L}^{(m)}_k$, over communication rounds for the benchmark and proposed schemes.}
\label{avg_loss}
\vspace{-0.1cm}
\end{figure}
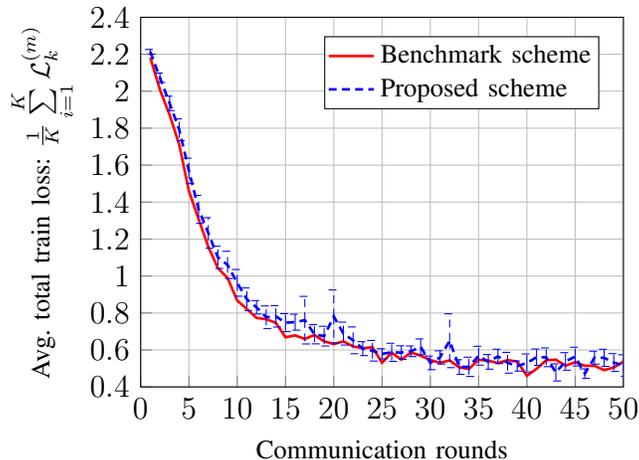

\subsection{Validation}

In this subsection, we first perform the comparative convergence analysis of the proposed scheme with the benchmark scheme (which represents existing state-of-the-art work). Fig.~\ref{avg_loss} plots the average MTD total train loss, $\frac{1}{K}\sum_{i=1}^{K} \mathcal{L}^{(m)}_k$, over the communication rounds for the benchmark and proposed schemes. The simulations were run for 200 rounds but only first 50 are shown in Fig.~\ref{avg_loss} for better clarity. Although the channel heterogeneity (including the channel realization and the path loss) exists among MTDs, in any given round, the average loss and the standard deviation in loss across different MTDs is very similar for both schemes. Specifically, the average training loss is only 6.26\% higher for the proposed scheme. This shows that the overall learning performance does not suffer from the proposed MTD-wise adaptive global model quality control. In addition, a small standard deviation, i.e., around 0.103, in training loss indicates that the learning experience is fair among MTDs.

\begin{figure}[t]
\centering
    \begin{tikzpicture}[spy using outlines={ chamfered rectangle, magnification=2.0, width=1.25cm, height=2.0cm,, connect spies}]
    \begin{axis}[
     height=6.5cm, width=8.0cm,
     legend cell align=left,
     legend style={inner xsep=1pt, inner ysep=1pt,at={(0.57,0.85)},anchor=east,font=\small, legend columns=1, draw, fill},
     cycle list name = mycyclelist,
     mark repeat={1},
     grid=both,
     label style={font=\small},
     xtick pos=left,
     ytick pos=left,
     xmin=0, xmax=1, xtick={0,0.1,0.2,0.3,0.4,0.5,0.6,0.7,0.8,0.9,1},
     xlabel= Normalised path loss of MTDs,
     ylabel= Avg. tot. energy: $\frac{1}{M}\sum\limits_{m=1}^{M} E^{(m)}_{\text{cp+tx},k}$ (mW),
     ylabel style={yshift=-0.5ex},
     ymin=175, ymax=375, ytick={175,200,225,250,275,300,325,350,375},
     ticklabel style={
        /pgf/number format/fixed,
        /pgf/number format/precision=5
     }
     ]
        \addplot table[x expr={\thisrow{x}/9-10/9+1}, y expr={\thisrow{y}*1000}] {avg_E_tot_b.txt};
        \addplot table[x expr={\thisrow{x}/9-10/9+1}, y expr={\thisrow{y}*1000}] {avg_E_tot_p.txt};
    \legend{Benchmark scheme, Proposed scheme}
    \end{axis}
        \end{tikzpicture}
\caption{The average total energy cost of $k$th~MTD, $\frac{1}{M}\sum_{m=1}^{M} E^{(m)}_{\text{cp+tx},k}$, per communication round versus its path loss for the benchmark and proposed schemes.}
\label{avg_E_tot}
\vspace{-0.1cm}
\end{figure}
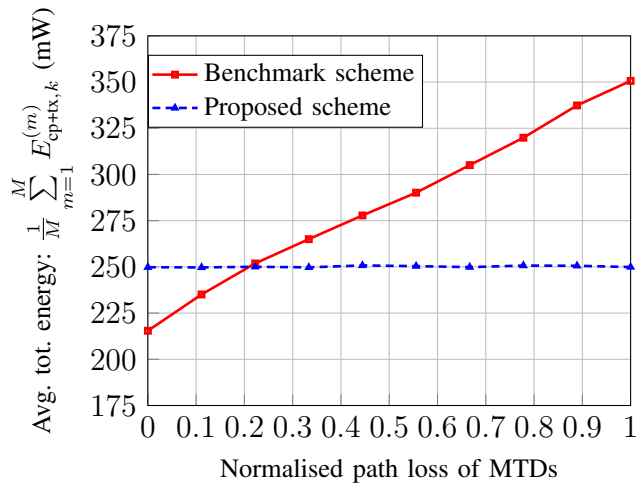

From Fig.~\ref{avg_loss}, we observed fairness in learning despite the divergent global model quality of MTDs. Now we analyze the energy expenditure of channel-heterogeneous MTDs for the same simulation setup. In that Fig.~\ref{avg_E_tot} plots the average total (sum of computation and transmission) energy cost of $k$th~MTD, $\frac{1}{M}\sum_{m=1}^{M} E^{(m)}_{\text{cp+tx},k}$, per communication round versus its path loss for the benchmark and proposed schemes. As expected, for the benchmark scheme the total energy cost significantly increases with the path loss severity due to the so-called near-far problem. It is because, in the vanilla FL setting the focus is kept on the local model quality and as many as possible iterations are performed for given delay bound. In contrast, for the proposed scheme the total energy cost remains almost flat for all MTDs irrespective of the channel statistics. Thereby, as desired, the MTDs spend similar energy to learn similar quality of the global model (utility fairness). Specifically, the proposed scheme reduces the standard deviation of the energy cost across MTDs by 99\% and provides about 12.17\% reduction in the average energy cost across MTDs as compared to the benchmark scheme. Importantly the proposed scheme achieves this utility fairness without any knowledge of the computation or transmission expenditure of MTDs at the AP or among MTDs.

\subsection{Performance Analysis}

\begin{figure}[t]
\begin{subfigure}{.5\textwidth}
    \begin{tikzpicture}[spy using outlines={ chamfered rectangle, magnification=2.0, width=1.25cm, height=2.0cm,, connect spies}]
    \begin{axis}[
     height=6.5cm, width=8.0cm,
     legend cell align=left,
     legend style={inner xsep=1pt, inner ysep=1pt,at={(0.95,0.85)},anchor=east,font=\small, legend columns=1, draw, fill},
     cycle list name = mycyclelist,
     mark repeat={1},
     grid=both,
     label style={font=\small},
     xtick pos=left,
     ytick pos=left,
     xmin=0, xmax=1, xtick={0,0.1,0.2,0.3,0.4,0.5,0.6,0.7,0.8,0.9,1},
     xlabel= Normalised path loss of MTDs,
     ylabel= Avg. comp. energy: $\frac{1}{M}\sum\limits_{m=1}^{M} E^{(m)}_{\text{cp},k}$ (mW),
     ylabel style={yshift=-0.5ex},
     ymin=46, ymax=58, ytick={46,48,50,52,54,56,58},
     ticklabel style={
        /pgf/number format/fixed,
        /pgf/number format/precision=5
     }
     ]
        \addplot table[x expr={\thisrow{x}/9-10/9+1}, y expr={\thisrow{y}*1000}] {avg_E_cp_b.txt};
        \addplot table[x expr={\thisrow{x}/9-10/9+1}, y expr={\thisrow{y}*1000}] {avg_E_cp_p.txt};
    \legend{Benchmark scheme, Proposed scheme}
    \end{axis}
        \end{tikzpicture}
\caption{Average computation energy cost.}
\label{avg_E_cp}
\end{subfigure}
\begin{subfigure}{.5\textwidth}
\vspace{0.35cm}
    \begin{tikzpicture}[spy using outlines={ chamfered rectangle, magnification=2.0, width=1.25cm, height=2.0cm,, connect spies}]
    \begin{axis}[
     height=6.5cm, width=8.0cm,
     legend cell align=left,
     legend style={inner xsep=1pt, inner ysep=1pt,at={(0.67,0.85)},anchor=east,font=\small, legend columns=1, draw, fill},
     cycle list name = mycyclelist,
     mark repeat={1},
     grid=both,
     label style={font=\small},
     xtick pos=left,
     ytick pos=left,
     xmin=0, xmax=1, xtick={0,0.1,0.2,0.3,0.4,0.5,0.6,0.7,0.8,0.9,1},
     xlabel= Normalised path loss of MTDs,
     ylabel= Avg. trans. energy: $\frac{1}{M}\sum\limits_{m=1}^{M} E^{(m)}_{\text{tx},k}$ (mW),
     ylabel style={yshift=-0.5ex},
     ymin=125, ymax=325, ytick={125,150,175,200,225,250,275,300,325},
     ticklabel style={
        /pgf/number format/fixed,
        /pgf/number format/precision=5
     }
     ]
        \addplot table[x expr={\thisrow{x}/9-10/9+1}, y expr={\thisrow{y}*1000}] {avg_E_tx_b.txt};
        \addplot table[x expr={\thisrow{x}/9-10/9+1}, y expr={\thisrow{y}*1000}] {avg_E_tx_p.txt};
    \legend{Benchmark scheme, Proposed scheme}
    \end{axis}
        \end{tikzpicture}
\caption{Average transmission energy cost.}
\label{avg_E_tx}
\end{subfigure}
\caption{The average computation energy cost of $k$th~MTD, $\frac{1}{M}\sum_{m=1}^{M} E^{(m)}_{\text{cp},k}$, and the average transmission energy cost of $k$th~MTD, $\frac{1}{M}\sum_{m=1}^{M} E^{(m)}_{\text{tx},k}$, per communication round versus its path loss for the benchmark and proposed schemes.}
\label{avg_E_cp_tx}
\vspace{-0.1cm}
\end{figure}
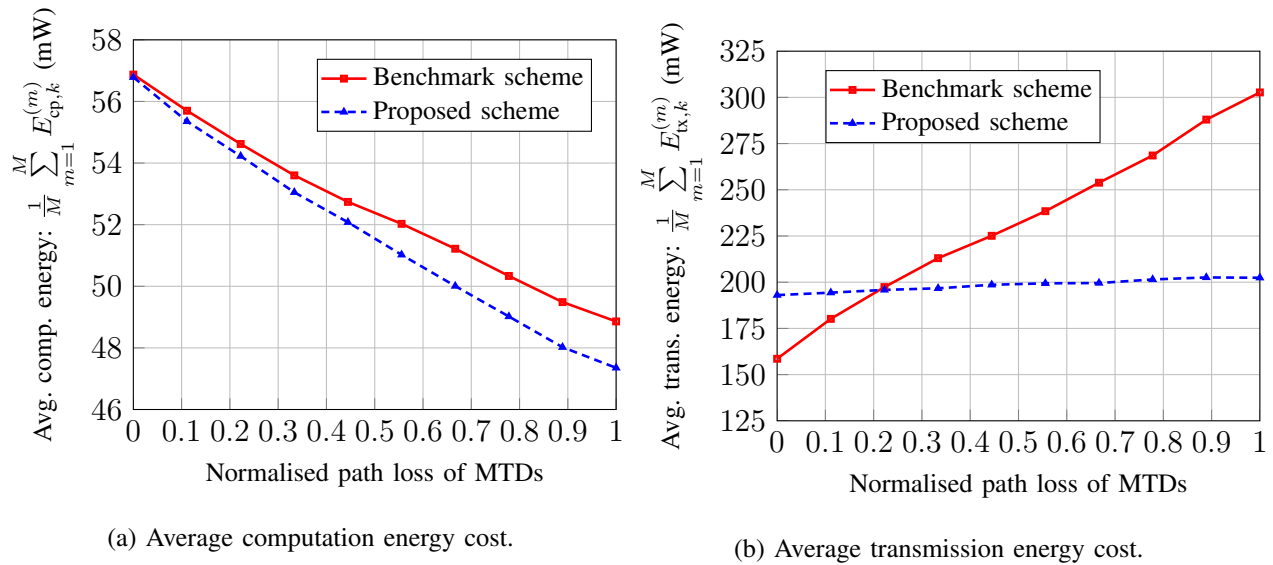

We know that for MTDs typically the computation energy cost is significantly less than the transmission energy cost. To analyse this, in Fig.~\ref{avg_E_cp_tx}(a) we plot the average computation energy cost of $k$-th~MTD, $\frac{1}{M}\sum_{m=1}^{M} E^{(m)}_{\text{cp},k}$, per communication round versus its path loss for the benchmark and proposed schemes. From Fig.~\ref{avg_E_cp_tx}(a), we can observe that the computation cost energy decreases with path loss severity for both schemes, however, the decrease is larger for the proposed scheme. Because the MTD with relatively poor channel experience is compelled to reduce the number of iterations to save time and energy, this in turn helps in reducing the transmission rate, thereby reducing the overall energy cost.

Fig.~\ref{avg_E_cp_tx}(b) plots the average transmission energy cost of $k$-th~MTD, $\frac{1}{M}\sum_{m=1}^{M} E^{(m)}_{\text{tx},k}$, per communication round versus its path loss for the benchmark and proposed schemes. In Fig.~\ref{avg_E_cp_tx}(b), we see that the transmission energy cost, in comparison to the benchmark scheme, increases only slightly for the proposed scheme with the MTD path loss severity. This is due to the adaptive local model quality control as per the proposed strategy. On the other hand, the benchmark scheme pushes each MTD irrespective of their path loss, and thus the transmission energy cost for far-off MTDs rises staggeringly. This anomaly results in severe unfairness issue among MTDs with heterogeneous channel experiences.

\begin{figure}[t]
\begin{subfigure}{.5\textwidth}
    \begin{tikzpicture}[spy using outlines={ chamfered rectangle, magnification=2.0, width=1.25cm, height=2.0cm,, connect spies}]
    \begin{axis}[
     height=6.5cm, width=8.0cm,
     legend cell align=left,
     legend style={inner xsep=1pt, inner ysep=1pt,at={(0.95,0.85)},anchor=east,font=\small, legend columns=1, draw, fill},
     cycle list name = mycyclelist,
     mark repeat={1},
     grid=both,
     label style={font=\small},
     xtick pos=left,
     ytick pos=left,
     xmin=0, xmax=1, xtick={0,0.1,0.2,0.3,0.4,0.5,0.6,0.7,0.8,0.9,1},
     xlabel= Normalised path loss of MTDs,
     ylabel= Avg. no. of iterations: $\frac{1}{M}\sum\limits_{m=1}^{M} j_k$ (mW),
     ylabel style={yshift=-0.5ex},
     ymin=80, ymax=100, ytick={80,82,84,86,88,90,92,94,96,98,100},
     ticklabel style={
        /pgf/number format/fixed,
        /pgf/number format/precision=5
     }
     ]
        \addplot table[x expr={\thisrow{x}/9-10/9+1}, y expr={\thisrow{y}*1}] {avg_iter_b.txt};
        \addplot table[x expr={\thisrow{x}/9-10/9+1}, y expr={\thisrow{y}*1}] {avg_iter_p.txt};
    \legend{Benchmark scheme, Proposed scheme}
    \end{axis}
        \end{tikzpicture}
\caption{Average no. of iterations performed by $k$-th~MTD.}
\label{avg_j}
\end{subfigure}
\begin{subfigure}{.5\textwidth}
\vspace{0.35cm}
    \begin{tikzpicture}[spy using outlines={ chamfered rectangle, magnification=2.0, width=1.25cm, height=2.0cm,, connect spies}]
    \begin{axis}[
     height=6.5cm, width=8.0cm,
     legend cell align=left,
     legend style={inner xsep=1pt, inner ysep=1pt,at={(0.95,0.85)},anchor=east,font=\small, legend columns=1, draw, fill},
     cycle list name = mycyclelist,
     mark repeat={1},
     grid=both,
     label style={font=\small},
     xtick pos=left,
     ytick pos=left,
     xmin=0, xmax=1, xtick={0,0.1,0.2,0.3,0.4,0.5,0.6,0.7,0.8,0.9,1},
     xlabel= Normalised path loss of MTDs,
     ylabel= Avg. trans. rate: $\frac{1}{M}\sum\limits_{m=1}^{M} R_k$ (Mbps),
     ylabel style={yshift=-0.5ex},
     ymin=3.50, ymax=5.75, ytick={3.50,3.75,4.0,4.25,4.50,4.75,5.0,5.25,5.50,5.75},
     ticklabel style={
        /pgf/number format/fixed,
        /pgf/number format/precision=5
     }
     ]
        \addplot table[x expr={\thisrow{x}/9-10/9+1},y expr={\thisrow{y}/1000000}] {avg_R_b.txt};
        \addplot table[x expr={\thisrow{x}/9-10/9+1},y expr={\thisrow{y}/1000000}] {avg_R_p.txt};
    \legend{Benchmark scheme, Proposed scheme}
    \end{axis}
        \end{tikzpicture}
\caption{Average transmission rate of $k$-th~MTD.}
\label{avg_R}
\end{subfigure}
\caption{The average no. of iterations performed by $k$-th~MTD, $\frac{1}{M}\sum_{m=1}^{M} j_k$, and the average transmission rate of $k$-th~MTD, $\frac{1}{M}\sum_{m=1}^{M} R_k$, per communication round versus its path loss for the benchmark and proposed schemes.}
\label{avg_j_R}
\vspace{-0.1cm}
\end{figure}
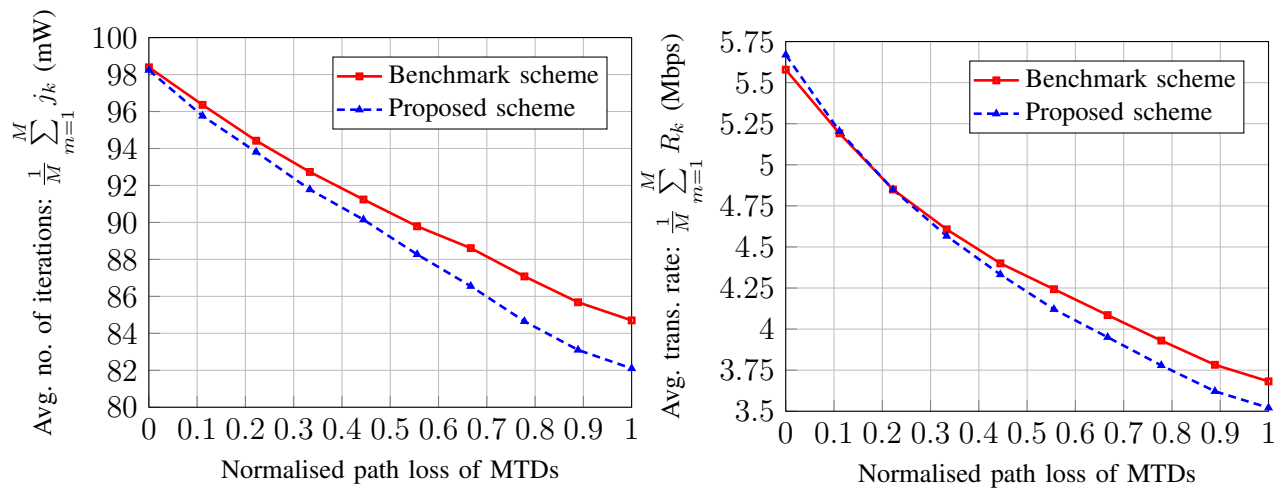

Let us now illustrate the impact of channel heterogeneity on the trade-off between the number of iterations that can be performed and the data transmission rate of the MTDs. Fig.~\ref{avg_j_R}(a) plots the average number of iterations performed by $k$-th~MTD, $\frac{1}{M}\sum_{m=1}^{M} j_k$, per communication round versus its path loss for the benchmark and proposed schemes. To satisfy the delay constraint both schemes strive to adapt the number of iterations and the transmission rate in each communication round. The proposed strategy takes another step to maintain similar average total energy cost among MTDs along similar learning gain. We can see from Fig.~\ref{avg_j_R}(a) that the number of iterations performed  decrease with the path loss severity. In particular, the MTDs with poor channel experience perform fewer iterations when proposed scheme is employed.

Fig.~\ref{avg_j_R}(b) plots the average transmission rate of $k$-th~MTD, $\frac{1}{M}\sum_{m=1}^{M} R_k$, per communication round versus its path loss for the benchmark and proposed schemes. As expected, we can observe from Fig.~\ref{avg_j_R}(b) that the transmission rate decreases with path loss severity. Interesting, in case of the proposed scheme, we can see that MTDs with good channel experience use a little higher transmission rate as compared to the benchmark scheme. Whereas, the transmission is much lower for the MTDs with poor channel experience.

\section{Conclusion}

In this paper, we have investigated the utility unfairness issue in a FL based wireless IoT network which is caused due to the device-heterogeneity. No prior work has jointly considered the diversity in devices' expenditure and contribution impacting the global model learning, and addressed the utility unfairness among devices.
We propose to control the quality of the global model shared with the devices, in each round, based on their contribution and expenditure. We design a utility function which works as a catalyst and it is used to reveal the optimal computation and transmission policies, such that the learning gain versus the cost is similar for all devices, without any knowledge of the contribution and expenditure of other devices.
Our results show that the proposed scheme reduces the standard deviation of the energy cost across MTDs by 99\% in comparison to the benchmark scheme, while the standard deviation of the training loss of MTDs varies around 0.103. The proposed scheme provides about 12.17\% reduction in the average energy cost across MTDs as compared to the benchmark scheme.
Here, we assumed that the instantaneous channel gain for each device is perfectly estimated by the AP. In a more practical case of imperfect channel estimation, outage occurs at the AP, and hence, an additional outage probability constraint needs to be introduced into the optimization problem. Nevertheless, the overall principle of the proposed scheme remains the same.

\appendices

\section{Proof of Theorem 1}\label{A}
From \eqref{assume-2} and \eqref{assume-3}, we have 
\begin{multline}\label{A-1}
    \mathcal{F}_k(\mathbf{w}^{(m)}_\textup{g},\mathbf{h}^{(m),(j+1)}_k) 
    \leq
    \mathcal{F}_k(\mathbf{w}^{(m)}_\textup{g},\mathbf{h}^{(m),(j)}_k) 
    + \Big(\mathbf{h}^{(m),(j+1)}_k - \mathbf{h}^{(m),(j)}_k \Big)^\intercal 
    \nabla \mathcal{F}_k(\mathbf{w}^{(m)}_\textup{g},\mathbf{h}^{(m),(j)}_k) 
    \\
    + \frac{L}{2} || \mathbf{h}^{(m),(j+1)}_k - \mathbf{h}^{(m),(j)}_k||^2.
\end{multline}

Substituting $\mathbf{h}^{(m),(j+1)}_k = \mathbf{h}^{(m),(j)}_k - \eta \nabla \mathcal{F}_k (\mathbf{w}^{(m)}_\textup{g}, \mathbf{h}^{(m),(j)}_k)$ from \eqref{loc-update-model} in \eqref{A-1}, we get
\begin{multline}\label{A-2}
    \mathcal{F}_k(\mathbf{w}^{(m)}_\textup{g},\mathbf{h}^{(m),(j+1)}_k) 
    \leq
    \mathcal{F}_k(\mathbf{w}^{(m)}_\textup{g},\mathbf{h}^{(m),(j)}_k) 
    - \eta
    || \nabla \mathcal{F}_k(\mathbf{w}^{(m)}_\textup{g},\mathbf{h}^{(m),(j)}_k) ||^2
    \\
    + \frac{L}{2} || - \eta \nabla \mathcal{F}_k(\mathbf{w}^{(m)}_\textup{g},\mathbf{h}^{(m),(j)}_k) ||^2.
\end{multline}

Rearranging \eqref{A-2} yields
\begin{equation}\label{A-3}
    \mathcal{F}_k(\mathbf{w}^{(m)}_\textup{g},\mathbf{h}^{(m),(j+1)}_k) 
    \leq
    \mathcal{F}_k(\mathbf{w}^{(m)}_\textup{g},\mathbf{h}^{(m),(j)}_k) 
    + \Big(\frac{\eta^2 L}{2} - \eta \Big) 
    || \nabla \mathcal{F}_k(\mathbf{w}^{(m)}_\textup{g},\mathbf{h}^{(m),(j)}_k) ||^2.
\end{equation}

Since, $\nabla \mathcal{F}_k(\mathbf{w}^{(m)}_\textup{g},\mathbf{h}^{*(m)}_k) = 0$ if $\mathbf{h}^{*(m)}_k$ is the optimal solution for \eqref{loc-opt-prob}, therefore we can write 
\begin{multline}\label{A-4}
|| \nabla \mathcal{F}_k(\mathbf{w}^{(m)}_\textup{g},\mathbf{h}^{(m),(j)}_k) ||^2 
= || \nabla \mathcal{F}_k(\mathbf{w}^{(m)}_\textup{g},\mathbf{h}^{(m),(j)}_k) - \nabla \mathcal{F}_k(\mathbf{w}^{(m)}_\textup{g},\mathbf{h}^{*(m)}_k) ||^2
\\
= || \nabla \mathcal{F}_k(\mathbf{w}^{(m)}_\textup{g},\mathbf{h}^{(m),(j)}_k) {-} \nabla \mathcal{F}_k(\mathbf{w}^{(m)}_\textup{g},\mathbf{h}^{*(m)}_k) ||
~
|| \nabla \mathcal{F}_k(\mathbf{w}^{(m)}_\textup{g},\mathbf{h}^{(m),(j)}_k) {-} \nabla \mathcal{F}_k(\mathbf{w}^{(m)}_\textup{g},\mathbf{h}^{*(m)}_k) ||.
\end{multline}

From \eqref{assume-1}, we have 
\begin{equation}\label{A-5}
    ||\nabla\mathcal{F}_k(\mathbf{w}^{(m)}_\textup{g},\mathbf{h}^{(m),(j)}_k) -
    \nabla\mathcal{F}_k(\mathbf{w}^{(m)}_\textup{g},\mathbf{h}^{*(m)}_k)|| 
    \leq
    L ||\mathbf{h}^{(m),(j)}_k - \mathbf{h}^{*(m)}_k||.
\end{equation}

Substituting $||\nabla\mathcal{F}_k(\mathbf{w}^{(m)}_\textup{g},\mathbf{h}^{(m),(j)}_k) - \nabla\mathcal{F}_k(\mathbf{w}^{(m)}_\textup{g},\mathbf{h}^{*(m)}_k)||$ from \eqref{A-5} in \eqref{A-4} yields
\begin{multline}\label{A-6}
|| \nabla \mathcal{F}_k(\mathbf{w}^{(m)}_\textup{g},\mathbf{h}^{(m),(j)}_k) ||^2 
\leq 
L \Big( \mathbf{h}^{(m),(j)}_k - \mathbf{h}^{*(m)}_k \Big)^\intercal
\\
\Big( \nabla \mathcal{F}_k(\mathbf{w}^{(m)}_\textup{g},\mathbf{h}^{(m),(j)}_k) - \nabla \mathcal{F}_k(\mathbf{w}^{(m)}_\textup{g},\mathbf{h}^{*(m)}_k) \Big).
\end{multline}

Since, $\nabla \mathcal{F}_k(\mathbf{w}^{(m)}_\textup{g},\mathbf{h}^{*(m)}_k) = 0$ this implies that
\begin{equation}\label{A-7}
|| \nabla \mathcal{F}_k(\mathbf{w}^{(m)}_\textup{g},\mathbf{h}^{(m),(j)}_k) ||^2
\leq
L \Big( \mathbf{h}^{(m),(j)}_k - \mathbf{h}^{*(m)}_k \Big)^\intercal
\nabla \mathcal{F}_k(\mathbf{w}^{(m)}_\textup{g},\mathbf{h}^{(m),(j)}_k).
\end{equation}

Also, from \eqref{assume-2} and \eqref{assume-3}, we have
\begin{multline}\label{A-8}
    \mathcal{F}_k(\mathbf{w}^{(m)}_\textup{g},\mathbf{h}^{*(m)}_k) 
    \leq
    \mathcal{F}_k(\mathbf{w}^{(m)}_\textup{g},\mathbf{h}^{(m),(j)}_k) 
    + \left( \mathbf{h}^{*(m)}_k - \mathbf{h}^{(m),(j)}_k \right)^\intercal 
    \nabla \mathcal{F}_k(\mathbf{w}^{(m)}_\textup{g},\mathbf{h}^{(m),(j)}_k) 
    \\
    + \frac{L}{2} || \mathbf{h}^{*(m)}_k - \mathbf{h}^{(m),(j)}_k||^2.
\end{multline}

Rearranging \eqref{A-8} yields
\begin{multline}\label{A-9}
    \mathcal{F}_k(\mathbf{w}^{(m)}_\textup{g},\mathbf{h}^{(m),(j)}_k) - \mathcal{F}_k(\mathbf{w}^{(m)}_\textup{g},\mathbf{h}^{*(m)}_k) 
    \geq
    \Big( \mathbf{h}^{(m),(j)}_k - \mathbf{h}^{*(m)}_k \Big)^\intercal
    \nabla \mathcal{F}_k(\mathbf{w}^{(m)}_\textup{g},\mathbf{h}^{(m),(j)}_k) 
    \\
    - \frac{L}{2} || \mathbf{h}^{*(m)}_k - \mathbf{h}^{(m),(j)}_k||^2.
\end{multline}

The bound on $\big( \mathbf{h}^{(m),(j)}_k - \mathbf{h}^{*(m)}_k \big)^\intercal \nabla \mathcal{F}_k(\mathbf{w}^{(m)}_\textup{g},\mathbf{h}^{(m),(j)}_k)$ in \eqref{A-9} allows us to rewrite \eqref{A-7} as
\begin{equation}\label{A-10}
|| \nabla \mathcal{F}_k(\mathbf{w}^{(m)}_\textup{g},\mathbf{h}^{(m),(j)}_k) ||^2 
\leq 
L \left( \mathcal{F}_k(\mathbf{w}^{(m)}_\textup{g},\mathbf{h}^{(m),(j)}_k) - \mathcal{F}_k(\mathbf{w}^{(m)}_\textup{g},\mathbf{h}^{*(m)}_k) \right).
\end{equation}

Substituting $|| \nabla \mathcal{F}_k(\mathbf{w}^{(m)}_\textup{g},\mathbf{h}^{(m),(j)}_k) ||^2 $ from \eqref{A-10} in \eqref{A-3}
\begin{multline}\label{A-11}
    \mathcal{F}_k(\mathbf{w}^{(m)}_\textup{g},\mathbf{h}^{(m),(j+1)}_k) 
    \leq
    \mathcal{F}_k(\mathbf{w}^{(m)}_\textup{g},\mathbf{h}^{(m),(j)}_k) 
    \\
    + \mu \Big(\frac{ \eta^2 L}{2} {-} \eta \Big) 
    \left(
    \mathcal{F}_k(\mathbf{w}^{(m)}_\textup{g},\mathbf{h}^{(m),(j)}_k) {-} \mathcal{F}_k(\mathbf{w}^{(m)}_\textup{g},\mathbf{h}^{*(m)}_k)
    \right).
\end{multline}

Subtracting $\mathcal{F}_k(\mathbf{w}^{(m)}_\textup{g},\mathbf{h}^{*(m)}_k)$ from both sides and rearranging yields
\begin{multline}\label{A-12}
    \mathcal{F}_k(\mathbf{w}^{(m)}_\textup{g},\mathbf{h}^{(m),(j+1)}_k) 
    - \mathcal{F}_k(\mathbf{w}^{(m)}_\textup{g},\mathbf{h}^{*(m)}_k)
    \\
    \leq
    L \Big( \frac{ \eta^2 L}{2} {-}  \eta  {+} \frac{1}{L} \Big) 
    \left(
    \mathcal{F}_k(\mathbf{w}^{(m)}_\textup{g},\mathbf{h}^{(m),(j)}_k) {-} \mathcal{F}_k(\mathbf{w}^{(m)}_\textup{g},\mathbf{h}^{*(m)}_k)
    \right).
\end{multline}

As per the result in \eqref{A-12} for any iteration the accuracy is at least $\big( \frac{\eta^2 L^2}{2} - \eta L  + 1  \big)$. Thus, we rewrite \eqref{A-12} as
\begin{multline}\label{A-13}
    \mathcal{F}_k(\mathbf{w}^{(m)}_\textup{g},\mathbf{h}^{(m),(j+1)}_k) 
    - \mathcal{F}_k(\mathbf{w}^{(m)}_\textup{g},\mathbf{h}^{*(m)}_k)
    \\
    \leq
    \Big( \frac{\eta^2 L^2}{2} - \eta L + 1 \Big)  ^{j+1}
    \left(
    \mathcal{F}_k(\mathbf{w}^{(m)}_\textup{g},\mathbf{0}) {-} \mathcal{F}_k(\mathbf{w}^{(m)}_\textup{g},\mathbf{h}^{*(m)}_k)
    \right).
\end{multline}

Rearranging \eqref{A-13} yields
\begin{equation}\label{A-14}
\frac{\mathcal{F}_k(\mathbf{w}^{(m)}_\textup{g},\mathbf{h}^{(m),(j+1)}_k)     - \mathcal{F}_k(\mathbf{w}^{(m)}_\textup{g},\mathbf{h}^{*(m)}_k)}
{\mathcal{F}_k(\mathbf{w}^{(m)}_\textup{g},\mathbf{0}) - \mathcal{F}_k(\mathbf{w}^{(m)}_\textup{g},\mathbf{h}^{*(m)}_k)}
    \leq
    \Big( \frac{\eta^2 L^2}{2} - \eta L  + 1 \Big) ^{j+1}.
    \hspace{-0.30cm}
\end{equation}

From \eqref{loc-accuracy}, if the $k$-th MTD solves the local problem in \eqref{loc-opt-prob} with accuracy $\Phi$, i.e., 
\begin{equation}\label{A-15}
    \Phi = \frac{\mathcal{F}_k(\mathbf{w}^{(m)}_\textup{g},\mathbf{h}^{(m),(j)}_k)-\mathcal{F}_k(\mathbf{w}^{(m)}_\textup{g},\mathbf{h}^{*(m)}_k)  } {\mathcal{F}_k(\mathbf{w}^{(m)}_\textup{g},\mathbf{0})-\mathcal{F}_k(\mathbf{w}^{(m)}_\textup{g},\mathbf{h}^{*(m)}_k)},
\end{equation}
\noindent then we can rewrite \eqref{A-14} as
\begin{equation}\label{A-16}
\Phi \leq \Big( \frac{\eta^2 L^2}{2} - \eta L  + 1 \Big) ^{j}.
\end{equation}

Finally, by solving \eqref{A-16} for $j$ yields \eqref{iter-bound}, i.e., the lower bound on the number of iterations required to achieve accuracy $\Phi$, the theorem now follows.

\section{Proof of Lemma 1}\label{B}

It can be shown that the transmission energy cost of the $k$-th MTD (which is part of $\mathcal{U}_k$), $E_{\textup{tx}, k} = T_{\textup{tx}, k} P_{\textup{tx}, k}$, is non-convex in $P_k$. By substitution of variable $Z_k = \log \big(  1 + \kappa \frac {P_k |h_k|^2}{\sigma^2_\textup{awgn} r^\alpha_k \Gamma } \big)$, transmission energy cost $E_{\textup{tx}, k}$ can equivalently be expressed as
\begin{equation}\label{E-tx-Z}
  E_{\textup{tx}, k} \big(Z_i \big) =  \frac{\mathcal{V}_v b_k} {Z_k}  \big( \exp  ( Z_k ) + c_k  \big),
\end{equation}
\noindent where $b_k=\frac{ \sigma^2_\textup{awgn} r^{\alpha}_k \Gamma \log(2)}{\rho \kappa B_k |h_k|^2}$, $c_k = \frac{\rho \kappa |h_k|^2 P_{\textup{cir},k} } { \sigma^2_\textup{awgn} r^{\alpha}_k \Gamma} - 1$.
Substituting $Z_k$ in \eqref{utility-max-opt-prob-b} and \eqref{utility-max-opt-prob-d} yields
\begin{subequations}
\begin{alignat}{3}
& j_k d_k \tau_k + \frac{\mathcal{V}_v \log(2)}  {B_k Z_k} \leq T, \label{utility-max-opt-prob-b-2}\\
& Z_{\textup{min},k} \leqslant Z_k \leqslant Z_{\textup{max},k}, 
\label{utility-max-opt-prob-d-2}
\end{alignat}
\end{subequations}
\noindent where $Z_{\textup{x},k} = \log \big(  1 +  \frac {\kappa P_{\textup{x},k} |h_k|^2}{\sigma^2_\textup{awgn} r^\alpha_k \Gamma } \big)$, $x \in \{\text{min, max\}}$. Thereby, $E_{\textup{tx}, k} \big(Z_i \big)$ in \eqref{E-tx-Z} is convex in $Z_k$ and constraint functions in \eqref{utility-max-opt-prob-b-2} and \eqref{utility-max-opt-prob-d-2}, respectively, are jointly convex in $j_k$ and $Z_k$.

Consider $\Psi_k$, a function of $j_k$, as
\begin{equation}\label{Psi-func}
\Psi_k (j_k) = - \beta_{1,k} \exp \Big( - \frac{1}{\beta_{2,k}}
j_k d_k \tau_k P_{\textup{cp},k} \Big) + \beta_{1,k}.
\end{equation}
\noindent It is easy to show that $\Psi_k$ is concave in $j_k$, i.e., $\Psi''_k \leq 0$.

Consider $\Omega_k$, a composition function of functions $H_k$ and $G_k$, is given as
\begin{equation}\label{Omega-func}
\Omega_k = H_k \circ G_k = H_k\big(G_k(j_k, Z_k)\big), 
\end{equation}
\noindent where
\begin{equation}\label{g-func}
G_k(j_k, Z_k) = j_k d_k \tau_k P_{\textup{cp},k} +  \frac{\mathcal{V}_v b_k} {Z_k}  \big( \exp  ( Z_k ) + c_k  \big),
\end{equation}
\begin{equation}\label{h-func}
H_k(y) = - y^2  + \varrho y.
\end{equation}
This implies
\begin{equation}\label{Omega-func-2}
\Omega_k(j_k, Z_k) = - G_k(j_k, Z_k)^2  + \varrho G_k(j_k, Z_k).
\end{equation}

Let us define the condition for the convexity or concavity of a composition function from composite rules given in \cite{boyd2004}.

\begin{definition}
The composite function $\Omega = H ~\circ~G$ is concave, i.e., 
\begin{equation}\label{Omega-func-3}
\Omega''(y) = H''(G(y))G'(y)^2 + H'(G(y))G''(y)^2 \leq 0,
\end{equation}
\noindent if $H$ is concave ($H''\leq 0$) and nonincreasing ($H'\leq 0$), and $G$ is convex ($G''\geq 0$). 
\end{definition}

It is easy to show that $H_k(y)$ in \eqref{h-func} is concave and nonincreasing. Similarly it can also be shown that $G_k(j_k, Z_k)$ is jointly convex in $j_k$ and $Z_k$. Then from Definition~5, $\Omega_k(j_k, Z_k)$ is jointly concave in $j_k$ and $Z_k$.
Since, $\mathcal{U}_k(j_k, Z_k) = \Psi_k (j_k) + \Omega_k(j_k, Z_k)$ is a sum of two concave functions, thus $\mathcal{U}_k$ in \eqref{MTD-utility} is concave \cite{boyd2004}. 
Hence, the problem in \eqref{utility-max-opt-prob} can equivalently be given as the following convex optimization problem
\begin{equation}\label{utility-max-opt-prob-convex-tmp}
\begin{aligned}
&  \underset{j_k, Z_k }    {\textup{minimize}}
& &  -\mathcal{U}_k\big(j_k, Z_k \big) \\
&  \textup{subject to}
& &  \eqref{utility-max-opt-prob-b-2}, \, \eqref{utility-max-opt-prob-d-2}, \, \eqref{utility-max-opt-prob-e}.
\end{aligned}
\end{equation}

\section{Proof of Theorem 2}\label{C}

Lagrangian function for \eqref{utility-max-opt-prob-convex} can be given as
\begin{multline}\label{lag-main}
\mathscr{L}(j_k, Z_k,\mathcal{Q})  =  
\beta_{1,k} \exp \Big( - \frac{1}{\beta_{2,k}}
j_k d_k \tau_k P_{\textup{cp},k} \Big) 
+ \Big(j_k d_k \tau_k P_{\textup{cp},k} +  \frac{\mathcal{V}_v b_k} {Z_k}  \big( \exp  ( Z_k ) + c_k  \big) \Big)^2 
\\
- \varrho j_k d_k \tau_k P_{\textup{cp},k} - \beta_{1,k} -  \frac{\varrho \mathcal{V}_v b_k} {Z_k}  \big( \exp  ( Z_k ) + c_k  \big) 
+ \Lambda_{1}  \Big( j_k d_k \tau_k + \frac{\mathcal{V}_v \log(2)}  {B_k Z_k} - T \Big)
+ \Lambda_{2} ( Z_{\textup{min},k} - Z_k) 
\\
+ \Lambda_{3} ( Z_k - Z_{\textup{max},k}) 
+ \Lambda_{4} ( j_{\textup{min},k} - j_k) 
+ \Lambda_{5} ( j_k - j_{\textup{max},k} ),
\end{multline}
\noindent where $\Lambda_i \in \mathcal{Q} = \{\Lambda_1, \Lambda_2, \Lambda_3, \Lambda_4, \Lambda_5\} $ is the Lagrangian multiplier associated with the $i$th constraint function. The Karush-Kuhn-Tucker (KKT) conditions for \eqref{utility-max-opt-prob-convex} are:
\begin{subequations}
\begin{equation}
\begin{aligned}
& j_k d_k \tau_k + \frac{\mathcal{V}_v \log(2)}  {B_k Z_k} - T \leqslant 0,  \; Z_{\textup{min},k} - Z_k \leqslant 0,
\\
& Z_k - Z_{\textup{max},k} \leqslant 0,  \; j_{\textup{min},k} - j_k \leqslant 0,
\; j_k - j_{\textup{max},k} \leqslant 0,
\end{aligned}
\end{equation}
\begin{equation}
\begin{aligned}
& \Lambda_{1} \geqslant 0, \Lambda_{2} \geqslant 0,  \Lambda_{3} \geqslant 0,  \Lambda_{4} \geqslant 0, \Lambda_{5} \geqslant 0, 
\end{aligned}
\end{equation}
\begin{equation}\label{3rd-kkt-1}
\begin{aligned}
& \Lambda_{1}  \Big( j_k d_k \tau_k + \frac{\mathcal{V}_v \log(2)}  {B_k Z_k} - T \Big) = 0, \; \Lambda_{2} (Z_{\textup{min},k} - Z_k) = 0,\\
& \Lambda_{3} ( Z_k {-} Z_{\textup{max},k}) = 0, 
\Lambda_{4} (j_{\textup{min},k} {-} j_k) = 0,  
\Lambda_{5} ( j_k {-} j_{\textup{max},k} ) = 0.\\
\end{aligned}
\end{equation}
\begin{equation}\label{4th-kkt-1}
\begin{aligned}
& \nabla \mathscr{L}(j_k, Z_k,\Lambda) = \bigg[\frac{\partial \mathscr{L}}{\partial j_k} ~ \frac{\partial \mathscr{L}}{\partial Z_k} \bigg]^\intercal = [0~0]^\intercal.
\end{aligned}
\end{equation}
\end{subequations}
\noindent where $\nabla$ is the gradient operator and $[\cdot]^\intercal$ is the transpose operator.

Taking partial derivative of \eqref{lag-main} w.r.t. $j_k$ and setting $\frac{\partial \mathscr{L}}{\partial j_k} = 0$ and after simplification we get
\begin{multline}\label{lag-eq-1}
\hspace{-0.05cm}
-\frac{\beta_{1,k}}{\beta_{2,k}} \exp \Big( - \frac{1}{\beta_{2,k}} j_k d_k \tau_k P_{\textup{cp},k}  \Big) - \varrho 
+ 2  \Big(j_k d_k \tau_k P_{\textup{cp},k} +  \frac{\mathcal{V}_v b_k} {Z_k}  \big( \exp  ( Z_k ) + c_k  \big) \Big)
\\
+ \frac{\Lambda_{1}}{P_{\textup{cp},k}} - \frac{\Lambda_{4} - \Lambda_{5}}{d_k \tau_k P_{\textup{cp},k}}   =  0.
\end{multline}
Taking partial derivative of \eqref{lag-main} w.r.t. $Z_k$ and setting $\frac{\partial \mathscr{L}}{\partial Z_k} = 0$ and after simplification we get
\begin{equation}\label{lag-eq-2}
2  \Big(j_k d_k \tau_k P_{\textup{cp},k} +  \frac{\mathcal{V}_v b_k} {Z_k}  
\big( \exp  ( Z_k ) + c_k  \big) \Big) - \varrho  
-  \frac{\Lambda_{1} \mathcal{V}_v b_k  \log(2) + Z^2 b_k B_k (\Lambda_{2}  - \Lambda_{3})}
{  B_k \mathcal{V}_v b^2_k \big( {(Z_k - 1) \exp( Z_k ) - c_k} \big)}  
 =  0.
\end{equation}

From complimentary slackness condition \eqref{3rd-kkt-1} we know either $\Lambda_{i}$ is zero or the associated constraint function is zero for any given $i$. Let us consider one of the possible cases that $\Lambda_{1}$ exists and $\Lambda_{2}, \Lambda_{3}, \Lambda_{4}, \Lambda_{5}$ do not exist. Accordingly, \eqref{lag-eq-1}, \eqref{lag-eq-2}, and \eqref{3rd-kkt-1} can be given as,
\begin{equation}\label{lag-eq-3}
-\frac{\beta_{1,k}}{\beta_{2,k}} \exp \Big( - \frac{1}{\beta_{2,k}}
j_k d_k \tau_k P_{\textup{cp},k} \Big) - \varrho  + \frac{\Lambda_{1}}{P_{\textup{cp},k}}
+ 2  \Big(j_k d_k \tau_k P_{\textup{cp},k} +  \frac{\mathcal{V}_v b_k} {Z_k}  \big( \exp  ( Z_k ) + c_k  \big) \Big) =  0,
\end{equation}
\begin{equation}\label{lag-eq-4}
2  \Big(j_k d_k \tau_k P_{\textup{cp},k} +  \frac{\mathcal{V}_v b_k} {Z_k}  \big( \exp  ( Z_k ) + c_k  \big) \Big) - \varrho  
-  \frac{\Lambda_{1}  \log(2) }
{  B_k b_k \big( {(Z_k - 1) \exp( Z_k ) - c_k} \big)}  
 =  0,
\end{equation}
\begin{equation}\label{lag-eq-5}
   j_k d_k \tau_k + \frac{\mathcal{V}_v \log(2)}  {B_k Z_k} - T = 0,
\end{equation}
\noindent respectively. Solving \eqref{lag-eq-4} for $\Lambda_{1}$ yields
\begin{equation}\label{lag-eq-6}
\Lambda_{1} = \Big( \frac {B_k b_k}{ \log(2)} \big( {(Z_k - 1) \exp(Z_k) - c_k \big)}  \Big) 
\Big( 2 j_k d_k \tau_k P_{\textup{cp},k} +  \frac{2 \mathcal{V}_v b_k} {Z_k}  \big( \exp  ( Z_k ) + c_k  \big) - \varrho  \Big).
\end{equation}
\noindent Solving \eqref{lag-eq-5} for $j_k$ yields
\begin{equation}\label{lag-eq-7}
   j_k = \frac{1}{d_k \tau_k} \Big( - \frac{\mathcal{V}_v \log(2)}  {B_k Z_k} + T \Big).
\end{equation}

Substitute $\Lambda_{1}$ and $j_k$ from \eqref{lag-eq-6} and \eqref{lag-eq-7}, respectively, in \eqref{lag-eq-3} yields
\begin{multline}\label{lag-eq-8}
\Big( 2 P_{\textup{cp},k} \Big( T - \frac{\mathcal{V}_v \log(2)}  {B_k Z_k}  \Big)
+  \frac{2 \mathcal{V}_v b_k} {Z_k}  \big( \exp  ( Z_k ) + c_k  \big) - \varrho  \Big)
\\
\Big( \frac {B_k b_k}{P_{\textup{cp},k} \log(2)} \big( {(Z_k - 1) \exp( Z_k ) - c_k} \big) + 1 \Big)
= \frac{\beta_{1,k}}{\beta_{2,k}} \exp \Big( \frac{P_{\textup{cp},k}}{\beta_{2,k}}
  \Big(  \frac{\mathcal{V}_v \log(2)}  {B_k Z_k} - T \Big) \Big).
\end{multline}

Numerically solving \eqref{lag-eq-8} for $Z_k$ yields its value $\hat{Z}_k$. Substituting this value of $Z_k$ in \eqref{lag-eq-7} and solving for $j_k$ yields its value $\hat{j}_k$. It can be shown that $\hat{Z}_k$ and $\hat{j}_k$ satisfy all the KKT conditions, specifically $\Lambda_{1} \geq 0$,  when all constraints in \eqref{utility-max-opt-prob-convex} are slack except the first constraint. Following similar steps it can be shown that all other cases violate one or more KKT conditions. Hence, the derived solution in \eqref{lag-eq-7} and \eqref{lag-eq-8} is the optimal solution for the optimization problem \eqref{utility-max-opt-prob-convex}.

\end{document}